%% file: main.tex
\documentclass[10pt,twocolumn,letterpaper]{article}

\usepackage[pagenumbers]{cvpr} % To force page numbers, e.g. for an arXiv version

\input{preamble}

\definecolor{cvprblue}{rgb}{0.21,0.49,0.74}
\usepackage[pagebackref,breaklinks,colorlinks,citecolor=cvprblue]{hyperref}

\def\paperID{7225} % *** Enter the Paper ID here
\def\confName{CVPR}
\def\confYear{2024}

\title{DiffusionAtlas: High-Fidelity Consistent Diffusion Video Editing}
\author{Shao-Yu Chang\\
Academia Sinica\\
{\tt\small shaoyuc3@iis.sinica.edu.tw}
\and
Hwann-Tzong Chen\\
National Tsing Hua University\\
{\tt\small htchen@cs.nthu.edu.tw}
\and
Tyng-Luh Liu\\
Academia Sinica\\
{\tt\small liutyng@iis.sinica.edu.tw}
}

\begin{document}
\twocolumn[{%
\renewcommand\twocolumn[1][]{#1}%
\maketitle
\begin{center}
    \centering
    \captionsetup{type=figure}
    \includegraphics[width=\textwidth, trim=4 4 4 4,clip]{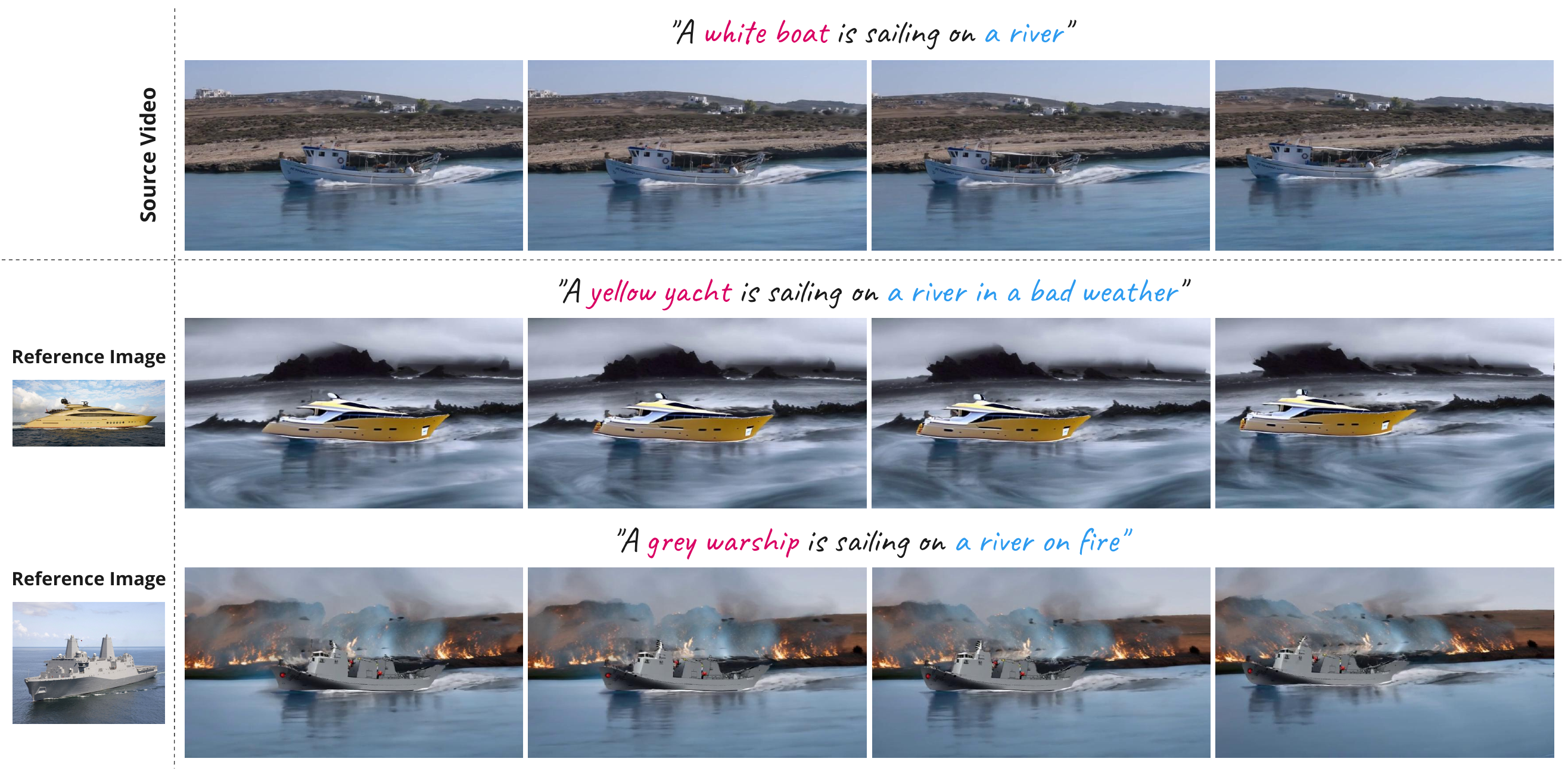}
    \captionof{figure}{\textbf{Video editing by DiffusionAtlas.} Given a source video, a text prompt, and a reference image, our method enables consistent video editing with high-fidelity output. (Words highlighted in red are foreground editings while blue are background editings.)}
    \label{fig:results_overview}
\end{center}%
}]
% Note that red is foreground, blue is background.

\input{sec/0_abstract}    
\input{sec/1_intro}
\input{sec/2_related}
\input{sec/3_method}
\input{sec/4_experiments}
\input{sec/5_limitations}
\input{sec/6_conclusion}

{
    \small
    \bibliographystyle{ieeenat_fullname}
    \bibliography{main}
}

\input{sec/X_supp_paper}

\end{document}

%% file: preamble.tex
\usepackage[dvipsnames]{xcolor}

\def\gD{{\mathcal{D}}}
\def\gE{{\mathcal{E}}}

\def\gI{{\mathcal{I}}}

\def\gL{{\mathcal{L}}}

\def\gN{{\mathcal{N}}}

\def\sA{{\mathbb{A}}}

\def\sM{{\mathbb{M}}}

%% file: sec/0_abstract.tex
\begin{abstract}
   We present a diffusion-based video editing framework, namely DiffusionAtlas, which can achieve both frame consistency and high fidelity in editing video object appearance. Despite the success in image editing, diffusion models still encounter significant hindrances when it comes to video editing due to the challenge of maintaining spatiotemporal consistency in the object's appearance across frames. On the other hand, atlas-based techniques allow propagating edits on the layered representations consistently back to frames. However, they often struggle to create editing effects that adhere correctly to the user-provided textual or visual conditions due to the limitation of editing the texture atlas on a fixed UV mapping field. Our method leverages a visual-textual diffusion model to edit objects directly on the diffusion atlases, ensuring coherent object identity across frames. We design a loss term with atlas-based constraints and build a pretrained text-driven diffusion model as pixel-wise guidance for refining shape distortions and correcting texture deviations. Qualitative and quantitative experiments show that our method outperforms state-of-the-art methods in achieving consistent high-fidelity video-object editing. Project page: \href{https://diffusionatlas.github.io/}{https://diffusionatlas.github.io/}.
   
   %Unlike image editing, diffusion-based methods still encounter significant limitations when it comes to video editing due to the challenge of maintaining temporal consistency over time. Existing works on layered representation of videos allow propagating edits consistently back to frames. However, they often struggle to wholly edit objects adherent properly to textual or visual conditions due to the limitation of editing on a fixed UV mapping field for texture atlas. Other works attempt to address this concern by applying a deformation formulation between objects. Yet shape distortion and pattern detail losses can often be seen in their results. In this paper, we propose a diffusion-based video editing framework, namely DiffusionAtlas, that is able to achieve both frame consistency and high-fidelity in object appearance. Our method leverages a visual-textual-based diffusion model to edit objects directly on the layered atlas textures, ensuring consistent object identity across frames. We then build up a loss term with atlas-based constraints and a pre-trained text-conditioned diffusion model as pixel-wise guidance for refining shape distortions and correcting texture shiftings. Qualitative and quantitative experiments demonstrate that our method outperforms state-of-the-art methods in achieving high-fidelity object video editing.
\end{abstract}

%% file: sec/1_intro.tex
\section{Introduction}
\label{sec:intro}
Diffusion-based generative models have made significant progress in enhancing the quality and diversity of generating or editing photorealistic images~\cite{kawar2023imagic, Kim_2022_CVPR, Rombach_2022_CVPR, brooks2022instructpix2pix, ruiz2023dreambooth, gal2022textual}. However, when it comes to video editing, it is a whole new level of challenge. Image generation methods often struggle to maintain temporal consistency between frames, as they do not consider the intricate dynamics of motion over time. On the other hand, naively modeling the entire video will involve huge computational complexity, which is not feasible for practical use. To address this issue, some approaches~\cite{wu2022tuneavideo, zhao2023makeaprotagonist} characterize only specific segments of video frames to reduce the computation cost. Yet, flickering artifacts and distorted shapes can often be seen in these methods as the information is insufficient to guide the models for generating new objects. Additional input guidance is thus considered to enhance temporal consistency and appearance coherence. Notably, the powerful tool ControlNet~\cite{zhang2023adding} has prompted vast research attention in video editing. Controlling the objects to align with extra conditions such as edge maps certainly improves spatiotemporal consistency and the fidelity of object appearances. However, these mechanisms hugely constrain the diversity and flexibility of editing objects, as they overlook the shape transformations between source and edited objects.
%Additionally, these mechanisms often struggle to produce videos with huge motions in it due to the still-existed flickering artifacts and unrealistic edited objects.

Layered Neural Atlases~\cite{kasten2021layered}  can decompose a video into unified appearance layers referred to as atlases, which facilitate consistent editing across individual frames with per-object UV space sampling association. While atlas-based methods~\cite{chai2023stablevideo, couairon2023videdit, bar2022text2live} have improved video quality and stabilized frame consistency, they are limited in manipulating object changes due to fixed UV space coordinates. Lee \etal~\cite{lee2023textvideoedit} present a deformation formulation to enable shape-aware editing. However, the shape deformation does not alter UV space structure and, consequently, sacrifices appearance details and disrupts object identities after editing.

This paper presents an atlas-based diffusion video editing approach: DiffusionAtlas. We utilize the fine-tuning strategy for a visual-textual-based video generative model to seek structure guidance from the input atlas, achieving reasonable object transformations that align well with query prompts. We also incorporate image embeddings from one of the video frames to control the content, allowing personalized reference images for object alterations during inference. Before the new temporal coherence is mapped back to the original frames, we optimize a new correlation between frame pixel locations and atlas sampling coordinates using constraints exploited from the original UV space to refine the object textures. Further, we adopt the gradient backpropagating procedure from DreamFusion~\cite{poole2022dreamfusion} to guide and revise the pixels to be generated in a more realistic manner.

We showcase the remarkable results of our method in achieving high-fidelity editing, preserving object identities and maintaining temporal consistency throughout frames without any additional semantic correspondence workarounds (see \cref{fig:results_overview}). Through extensive qualitative and quantitative experiments, our approach demonstrates superior results compared to state-of-the-art methods (see \cref{sec:experiments_4}). In summary, our contributions are as follows:

\begin{itemize}
    \item We present DiffusionAtlas, a high-fidelity video editing framework that achieves temporal consistency and object identity preservation for textual-visual-based editing.
    \item Our one-shot tuning mechanism for diffusion models can directly edit objects on the atlas textures, preserving input structures without using extra semantic correspondences.
    \item We conduct extensive experiments to show superior editing results compared to state-of-the-art methods.
\end{itemize}

%% file: sec/2_related.tex
\section{Related work} % textbf space adjustment
\label{sec:related}

\setlength{\parindent}{0pt}
\textbf{Image synthesis and editing.} Prior to the emergence of diffusion models~\cite{song2020denoising, ho2020denoising, NEURIPS2021_49ad23d1}, Generative Adversarial Networks (GANs)~\cite{li2020manigan, Patashnik_2021_ICCV, gal2021stylegannada, objgan19, liao2021text, pmlr-v139-ramesh21a, pmlr-v48-reed16, Tao18attngan} have dominated the area of generating visual contents. However, text-driven GAN-based methods are restricted to specific image domains and fall short of exhibiting text diversity. GLIDE~\cite{pmlr-v162-nichol22a} introduces classifier-free guidance~\cite{2022arXiv220712598H} to text-guided diffusion models,  improving the output quality while supporting image inpainting. DALLE-2~\cite{ramesh2022}, on the other hand, leverages the joint embedding space of CLIP~\cite{Radford2021LearningTV} to enrich image diversity and enhance caption similarity. Stable Diffusion~\cite{Rombach_2022_CVPR} enables training on limited computational resources by performing the denoising diffusion process in latent space. Other diffusion-based models have shown notable success in both text-guided image generation~\cite{gal2022textual, ruiz2023dreambooth, hu2022lora} and editing~\cite{meng2022sdedit, hertz2022prompt, Tumanyan_2023_CVPR, brooks2022instructpix2pix, kawar2023imagic} tasks. Our work leverages the state-of-the-art text-to-image model, Stable Diffusion~\cite{Rombach_2022_CVPR}, with the inversion of CLIP image embeddings~\cite{ramesh2022}, and extends the semantic image editing capabilities to consistent video editing.
%Recently, DiTs~\cite{Peebles2022DiT} inherits the scaling properties of transformer to diffusion models that operates on latent patches and has outperformed prior U-Net structures. Our work leverages the state-of-the-art text-to-image model, Stable Diffusion~\cite{Rombach_2022_CVPR}, with the inversion of CLIP image embeddings~\cite{ramesh2022}, and extends the semantic image editing capabilities to consistent video editing.

\begin{figure*}[htp!]
    \centering
    \includegraphics[width=\textwidth, trim=4 4 4 4,clip]{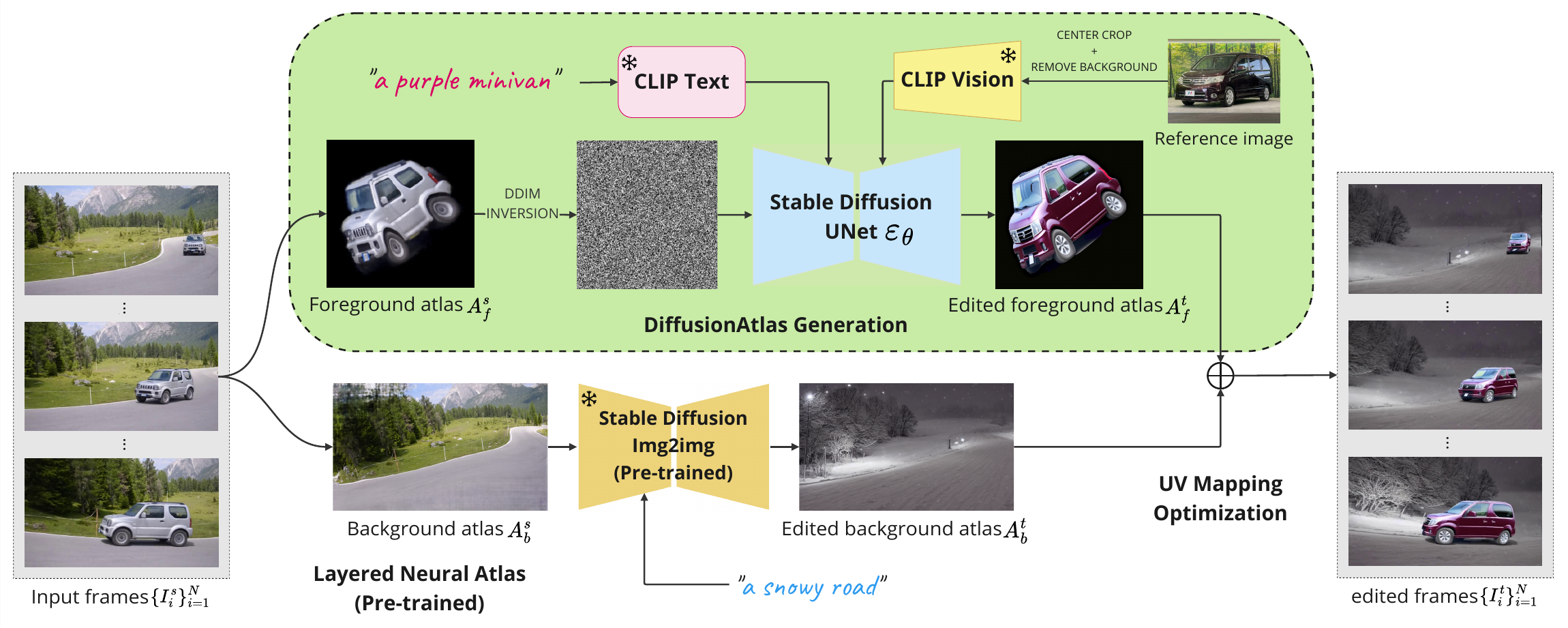}
    \caption{\textbf{Method overview.} The source video $\{I^s_i\}_{i=1}^N$ is first fed into a pretrained LNA~\cite{kasten2021layered} to obtain the foreground atlas $A^s_f$ and background atlas $A^s_b$. We edit the foreground and background in separate ways. For background editing, we leverage a pretrained image-to-image Stable Diffusion model~\cite{meng2022sdedit} to generate the target background atlas $A^t_b$ with the text prompt (in this example, the text prompt for background editing is ``a snowy road"). For foreground editing, we perform our DiffusionAtlas Generation method (\cref{sec:method_3.2}) on the foreground atlas $A^s_f$ with visual and textual clues. Note that we center-crop and remove the background of the reference image before feeding it into the CLIP Vision model. After getting the edited foreground atlas $A^t_f$, we optimize the UV mappings using constraints exploited from the original UV space structures and a pretrained diffusion model to propagate $A^t_f$ textures back to the frames and obtain edited frames $\{I^t_i\}_{i=1}^N$  (\cref{sec:method_3.3}).}
    \label{fig:model_pipeline}
\end{figure*}
%(visual embedding is obtained using CLIP Vision Model~\cite{Radford2021LearningTV} while textual cue is obtained using CLIP Text Model~\cite{Radford2021LearningTV})

\textbf{Diffusion for video generation and editing.} In contrast to the breakthrough in image editing, the research of video editing and generation still faces two core challenges: 1) maintaining temporal consistency and 2) managing the increased computational complexity of the additional dimension. Early text-to-video (T2V) generation works~\cite{ho2022video, imagen_video, singer2023makeavideo, 2022arXiv221111018Z} have shown impressive results. However, they suffer from expensive and complex computations due to heavily relying on large-scale image and video datasets. Tune-A-Video~\cite{wu2022tuneavideo} introduces a sparse spatiotemporal attention mechanism, considering only parts of the video frames to generate new content. However, since temporal correlations are not fully incorporated, geometry inconsistency and flickering artifacts can be seen in the edited videos. DreamMix~\cite{molad2023dreamix} enables text-based motion and appearance editing by jointly fine-tuning with full temporal attention and with temporal attention masking. Video-P2P~\cite{liu2023videop2p} and FateZero~\cite{qi2023fatezero} control intermediate attention maps during inversion to extend Prompt-to-prompt~\cite{hertz2022prompt} into video level. Make-A-Protagonist~\cite{zhao2023makeaprotagonist} uses textual and visual cues to provide coarse personalizations. Lately, ControlNet~\cite{zhang2023adding} has prompted many diffusion-based video synthesis methods~\cite{zhao2023controlvideo, 2023videocomposer, xing2023make} to focus on adding extra spatial controls to enhance temporal consistency. Yet, the conditioning controls restrict the diversity and degrade object fidelity that is supposed to reflect the text prompts for editing. 
%Unlike the methods mentioned above, our video editing framework can accommodate textual and visual information while maintaining high fidelity in object editing and preserving strong temporal consistency.

\textbf{Consistent video editing.} To attain temporally consistent editing effects, EBSynth~\cite{Jamriska19-SIG} takes keyframes and propagates edits to the entire video using optical flow as positional and temporal guidance. Omnimattes~\cite{lu2020, lu2021, lu2022} estimate RGBA layers for the target subject and scene effects for each frame independently, while Layered Neural Atlases (LNA)~\cite{kasten2021layered} can decompose objects and backgrounds into 2D layers for unified appearance representations. Text2Live~\cite{bar2022text2live} leverages a CLIP model~\cite{Radford2021LearningTV} to generate an edit layer that is composited over the original input. StableVideo~\cite{chai2023stablevideo} develops an inter-frame propagation mechanism using diffusion models and propagates the appearance information back to the video using the concept of object layered representations~\cite{kasten2021layered}.
On the other hand, VidEdit~\cite{couairon2023videdit} utilizes diffusion models on the 2D layered textures to edit object appearance before mapping back to frames. However, due to the fixed UV space mappings, these works can only allow appearance edits and are limited to object changes and editing diversity. Lee \etal~\cite{lee2023textvideoedit} achieves shape-aware consistency on edited objects by propagating the deformation field between input and edited keyframe to the whole video but suffers from object shape distortion and identity disruption. In contrast, our method can directly edit the atlas textures while achieving high-fidelity object changes that adhere properly to the input conditions without shape distortions or detail losses.

%% file: sec/3_method.tex
\section{DiffusionAtlas}
\label{sec:method}
When given an input video, a text prompt, and a reference image, our method is designed to generate an edited video in which the target object maintains both spatiotemporal consistency and high fidelity (see \cref{fig:model_pipeline} for our method overview). To ensure that no essential pixel details are lost during the propagation process, we utilize diffusion models directly on the 2D layered atlas textures, which are obtained from the pretrained video decomposition method, Layered Neural Atlases~\cite{kasten2021layered}. However, the fixed UV mappings can potentially lead to object distortions if we simply propagate the new atlas back to the video (see \cref{sec:ablation_study_4.3}). To address this, we employ an optimization process to create a new coordinate correlation between frames and the UV space, allowing us to propagate the edited atlas back to frames more naturally. This optimization procedure takes into account properties exploited from the original UV space structure. We also incorporate a pixel-level guidance score from DreamFusion~\cite{poole2022dreamfusion} to guide the network parameters.

\subsection{Preliminaries: layered neural atlases (LNA)}
\label{sec:method_3.1}
LNA decomposes a source video $\{I^s_i\}_{i=1}^N$ into unified 2D layered representations of the foreground object and the background. Each pixel location $p = (x, y, k) \in \mathbb{R}^{3}$ is fed into three mapping networks:
\begin{equation} \label{eq:nla_mappings} \textcolor{green}{\mathbb{M}_{f}(p)}
    \footnotesize
    \sM_{b}(p) = (u^{p,s}_b, v^{p,s}_b), \: \sM_{f}(p) = (u^{p,s}_f, v^{p,s}_f), \: \sM_{\alpha}(p) = {\alpha}^{p,s} \,,
\end{equation}
where $\sM_{b}$ and $\sM_{f}$ map $p$ to $(u, v)$-coordinate in the foreground and background atlas regions, respectively, and $\sM_{\alpha}$ predicts an opacity value for the visibility of $p$.
These $(u, v)$ coordinates are then fed into an atlas MLP $\sA$, which outputs the RGB color of $p$ at foreground and background atlases:
\begin{equation} \label{eq:nla_predict_rgb}
    c^{p,s}_b = \sA(u^{p,s}_b, v^{p,s}_b), \: c^{p,s}_f = \sA(u^{p,s}_f, v^{p,s}_f) \,.
\end{equation}
The final RGB color is then reconstructed by alpha-blending the predicted atlas points:
\begin{equation} \label{eq:nla_final_rgb}
    c^{p,s} = (1-{\alpha}^{p,s})c^{p,s}_b + {\alpha}^{p,s} c^{p,s}_f \,.
\end{equation}

Our method directly edits the RGB pixel values on the layered atlases using a Latent Diffusion Model (LDM)~\cite{Rombach_2022_CVPR}. We then optimize the $(u, v)$ maps to adopt the new atlas back to the original video frame coordinates.

\subsection{DiffusionAtlas generation}
\label{sec:method_3.2}
Recent atlas-based video editing methods rely on the 2D layered atlases to maintain temporal consistency during editing. However, none of them tries to edit objects directly in the space of foreground atlases, which is due to the warped object texture in the foreground atlas. The pretrained diffusion models are not trained for this kind of unnaturally distorted texture, as described by Lee \etal~\cite{lee2023textvideoedit}. VidEdit~\cite{couairon2023videdit} edits the atlas textures, but they constrain the object structures with segmentation masks and edge maps in order to achieve object appearance consistency. Lee \etal~\cite{lee2023textvideoedit} consider object shape awareness by finding semantic correspondences between the source object and target but suffer from object distortion and detail losses. This section demonstrates how our diffusion atlas editing method edits foreground atlases without disrupting object identities.

\textbf{Latent Diffusion Model with image embedding.} Our method leverages an LDM~\cite{Rombach_2022_CVPR} with image embedding~\cite{Radford2021LearningTV} to edit the atlas textures. LDMs consist of two main components. First, an autoencoder is trained to encode images $x$ into latent representations $z = \gE(x)$ with the encoder $\gE$ and reconstruct the latent back to pixel space with the decoder $\gD$, namely $\gD(\gE(x)) \approx x$. Then a DDPM~\cite{ho2020denoising} tries to remove the noise added to the sampled data in the latent space using a U-Net ${\varepsilon}_{\theta}$:
\begin{equation} \label{eq:noise_mse}
    \small
    \mathbb{E}_{z_0,\varepsilon \sim \gN(0,1),t,c,\gI}[||\sqrt{\alpha_t} \varepsilon - \sqrt{1-\alpha_t}z_0 - {\varepsilon}_{\theta}(z_t, t, c, \gI)||_2^2] \,,
\end{equation}
where $\alpha_t$ is the noise scheduler weight at each timestep $t$, and $c$ and $\gI$ denote the embedding of the text and image, respectively. Note that our method uses v-prediction~\cite{salimans2022progressive} instead of directly predicting the noise $\varepsilon$ as in~\cite{ho2020denoising} to improve the color consistency of videos. The text serves as the key and value for cross-attention~\cite{Rombach_2022_CVPR} while the image condition is directly added to features of residual blocks~\cite{ramesh2022}.

\begin{figure}[t!]
    \centering
    \includegraphics[width=0.47\textwidth,trim=4 4 4 4,clip]{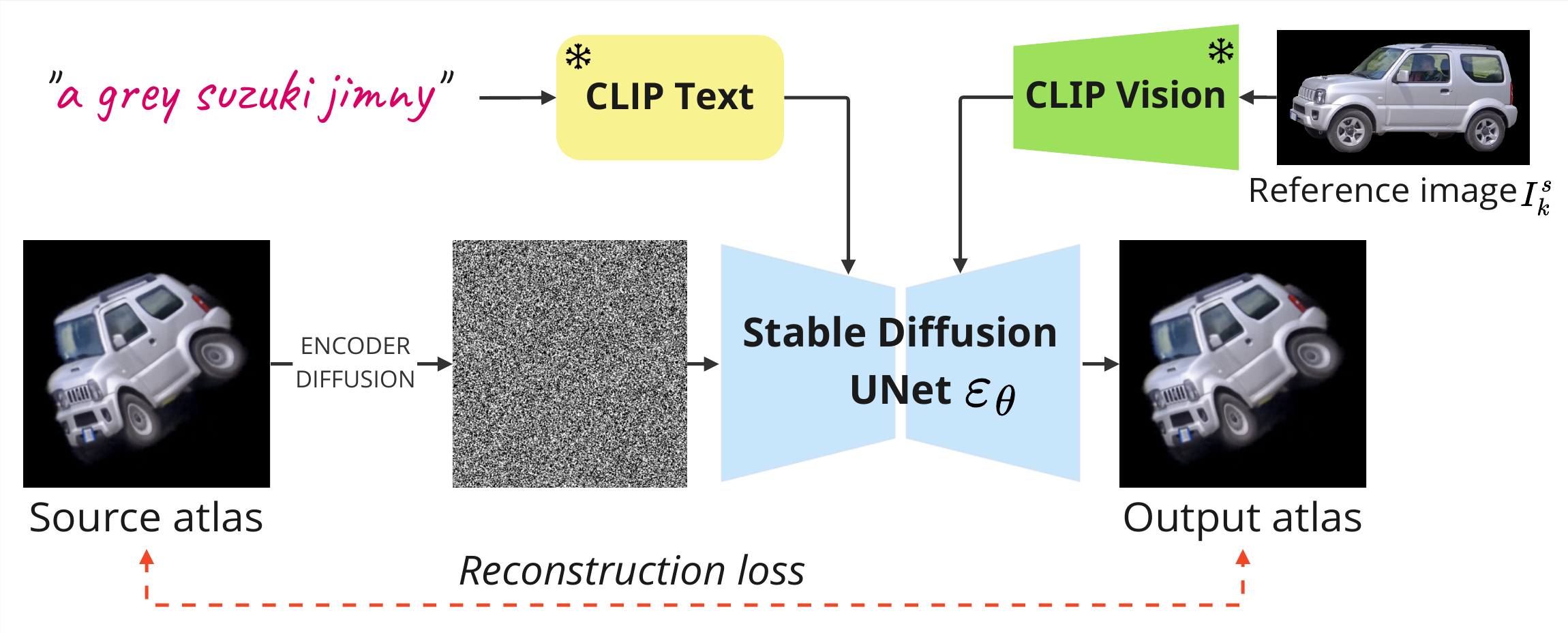}
    \caption{\textbf{Model fine-tuning process.} Given a source atlas input, our model is fine-tuned with the source caption and a randomly sampled frame $I^s_k$ from the input video as the reference image.} %Note that we center-crop the object and leave the background of the reference image black for more efficient training.
    \label{fig:diffusionAtlas_training}
\end{figure}

\begin{figure*}[t!]
    \centering
    \includegraphics[width=0.9\textwidth,trim=4 4 4 4,clip]{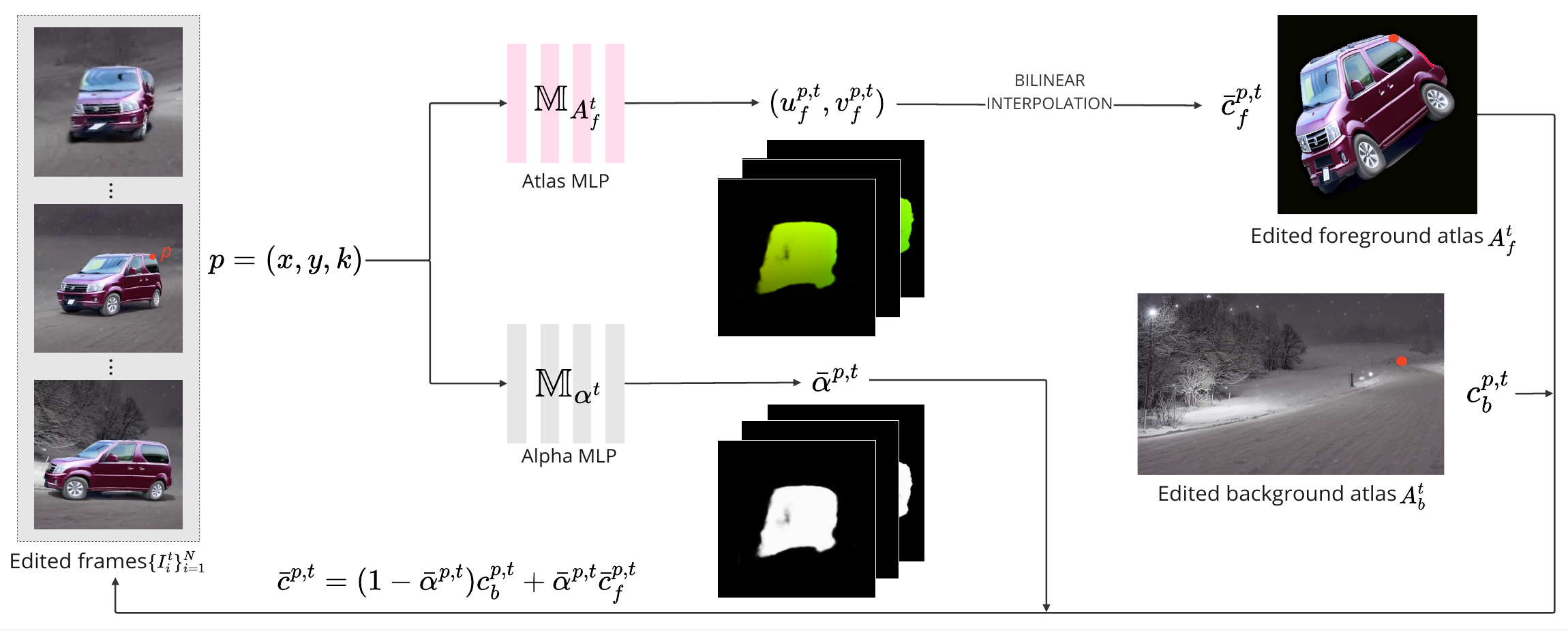}
    \caption{\textbf{UV mapping optimization process.} We optimize the UV mappings by feeding each pixel location $p = (x,y,k)$ of the edited frames $\{I^t_i\}_{i=1}^N$ into $\sM_{A_{f}^{t}}$ and $\sM_{{\alpha}^t}$. $\sM_{A_{f}^{t}}$ outputs a $(u,v)$ coordinate that is used to find the corresponding RGB value $\Bar{c}^{p,t}_f$ from the edited foreground atlas $A^t_f$ while $\sM_{{\alpha}^t}$ outputs an opacity value ${\Bar{\alpha}}^{p,t}$. We can then obtain the optimized edited RGB value $\Bar{c}^{p,t}$ for each $p$.}
    \label{fig:new_atlas_network}
\end{figure*}

\textbf{Model fine-tuning.} Following~\cite{singer2023makeavideo, 2022arXiv221111018Z, wu2022tuneavideo, zhao2023makeaprotagonist}, we fine-tune our network to update the attention matrix for the distorted source atlas. Unlike previous T2V generative models, we do not modify the U-Net structure as they are already implemented for 2D image generation. \cref{fig:diffusionAtlas_training} illustrates the fine-tuning process of our atlas generation procedure. Our atlas editing model takes into the input atlas, a respective caption, and a visual clue of the atlas object as conditions. The caption is transformed into a text embedding using CLIP Text Model~\cite{Radford2021LearningTV} while the visual embedding is encoded using CLIP Vision Model~\cite{Radford2021LearningTV}. To ensure the capability of incorporating image embedding, we randomly select one frame $I^s_k$ from the input video as the reference image at each iteration. Here, we center-crop the object within the frames and remove the background before feeding it into our model to help focus on the object and output an edited atlas with a black background.
%Note that text and image embeddings are applied to all the blocks in the U-Net.

\textbf{DDIM inversion.} We use deterministic DDIM sampling~\cite{song2020denoising} to generate images from latent noises in $T$ denoising steps:
\begin{equation} \label{eq:ddim_sampling}
    \footnotesize
    z_{t-1} = \sqrt{\frac{\alpha_{t-1}}{\alpha_t} }z_t + \sqrt{\alpha_{t-1}} \left(\sqrt{\frac{1}{\alpha_{t-1}} - 1} - \sqrt{\frac{1}{\alpha_{t}} - 1} \right) \cdot {\varepsilon}_{\theta} \,,
\end{equation}
where $t: 1 \rightarrow T$ denotes the timestamp, and $\alpha_t$ is the parameter for noise scheduling.

In order to maintain the distribution of the input structures as in fine-tuning, we implement DDIM inversion~\cite{song2020denoising} at inference to control pixel shifts, as described in~\cite{wu2022tuneavideo, zhao2023makeaprotagonist}. DDIM inversion transforms the input atlas into an initial latent noise, illustrated as follows:
\begin{equation} \label{eq:ddim_inversion}
    \footnotesize
    z_{t+1} = \sqrt{ \frac{\alpha_{t+1}}{\alpha_t} }z_t + \sqrt{\alpha_{t+1}} \left(\sqrt{\frac{1}{\alpha_{t+1}} - 1} - \sqrt{\frac{1}{\alpha_{t}} - 1} \right) \cdot {\varepsilon}_{\theta} \,.
\end{equation}

The target caption and reference image are then incorporated into the model to generate the target atlas (see \cref{fig:model_pipeline}).

\subsection{UV mapping optimization}
\label{sec:method_3.3}
After obtaining the new edited atlases $A_{f}^{t}$ and $A_{b}^{t}$ from the diffusion process in \cref{sec:method_3.2}, we are able to get an initial edited video by mapping the atlas texture back to frames using the pretrained LNA networks. Here, we apply Grounded-SAM~\cite{kirillov2023segany, liu2023grounding} on the edited atlas $A_{f}^{t}$ to get an initial alpha value ${\alpha}^{p,t}$ for each pixel location $p(x, y, k)$. Then, an initial edited RGB value for each $p$ can be obtained:
\begin{equation} \label{eq:initial_rgb}
    c^{p,t} = (1-{\alpha}^{p,t})c^{p,t}_b + {\alpha}^{p,t} c^{p,t}_f \,,
\end{equation}
where $c^{p,t}_f$ and $c^{p,t}_b$ are obtained via bilinear interpolation using the edited atlas textures $A_{f}^{t}$ and $A_{b}^{t}$ with the original pretrained MLP networks $\sM_{f}$ and $\sM_{b}$, respectively.

However, since we do not modify the original UV mappings, artifacts and shape distortions are shown in some areas of the rendered foreground object (see \cref{fig:ablation_study(b)} and \cref{sec:ablation_study_4.3}). To address this issue, we train an additional foreground atlas network $\sM_{A_{f}^{t}}$ and alpha network $\sM_{{\alpha}^t}$ to map the edited atlas back to the frame coordinates more naturally. Both networks are coordinate-based Multilayer Perceptron (MLP) representations for mappings and alphas, exploited from~\cite{kasten2021layered}. Here, each $p$ in frames is fed into both networks $\sM_{A_{f}^{t}}$ and $\sM_{{\alpha}^t}$. $\sM_{A_{f}^{t}}$ predicts the corresponding 2D coordinate to map each $p$ to the edited atlas $A_{f}^{t}$ while $\sM_{{\alpha}^t}$ determines the opacity value for $p$:
\begin{equation} \label{eq:new_atlas_mlp}
    \sM_{A_{f}^{t}}(p) = (u^{p,t}_f, v^{p,t}_f), \: \sM_{{\alpha}^t}(p) = {\Bar{\alpha}}^{p,t} \,.
\end{equation}
We can then obtain revisions of edited RGB values using the optimized UV mappings and opacity values:
\begin{equation} \label{eq:new_rgb}
    \Bar{c}^{p,t} = (1-{\Bar{\alpha}}^{p,t})c^{p,t}_b + {\Bar{\alpha}}^{p,t} \Bar{c}^{p,t}_f \,,
\end{equation}
where $\Bar{c}^{p,t}_f = A_{f}^{t}(u^{p,t}_f, v^{p,t}_f)$ is the foreground RGB value obtained from the edited atlas $A_{f}^{t}$ via bilinear interpolation using the predicted $(u,v)$ coordinate. Note that we center-crop the object in frames and focus only on the pixel locations that are within the cropped areas for more efficient training and to preserve temporal consistency in non-object areas. The UV mapping optimization process is shown in \cref{fig:new_atlas_network}.

\textbf{Loss terms.} We use two reconstruction losses to constrain how the optimized $(u,v)$ coordinates map back to frames, namely RGB and alpha reconstruction losses. Setting the initial edited videos as the starting point for our optimization, the RGB reconstruction loss measures the error between the initial edited RGB value $c^{p,t}$ from \cref{eq:initial_rgb} and the reconstructed RGB value from \cref{eq:new_rgb} by squared distance:
\begin{equation} \label{eq:rgb_recon_loss}
    \gL_\mathrm{rgb} = \|c^{p,t} - \Bar{c}^{p,t}\|_2^2 \,,
\end{equation}
and alpha reconstruction loss is the error between the mask obtained from Grounded-SAM~\cite{kirillov2023segany, liu2023grounding} and ${\Bar{\alpha}}^{p,t}$ obtained from \cref{eq:new_atlas_mlp}:
\begin{equation} \label{eq:alpha_loss}
    \gL_{\alpha} = \|{\alpha}^{p,t} - {\Bar{\alpha}}^{p,t}\|_2^2 \,.
\end{equation}

We also apply a coordinate structure loss to guide how the object structures should be formed. We encourage the offsets $\Delta$ between point $p(x, y, k)$ and $p'(x+\Delta, y+\Delta, k)$ to match the corresponding offsets on the optimized UV space according to the original UV mapping structure:
\begin{equation} \label{eq:corr_loss}
    \small
    \gL_\mathrm{off} = \|[\sM_{A_{f}^{t}}(p) - \sM_{A_{f}^{t}}(p')] - [\sM_{f}(p) - \sM_{f}(p')]\|_2^2 \,.
\end{equation}
This offset loss helps the networks connect points in the accurate directions and magnitudes (see supplementary for more detail).
%between foreground and non-foreground areas
%Note that we apply this loss not only in the non-foreground areas but throughout the whole atlas. We find out that this loss can help the networks build the object structures more efficiently since this property can connect points in accurate directions and magnitudes between foreground and non-foreground points.

\begin{figure*}[htp!]
    \centering
    \includegraphics[width=\textwidth, trim=4 4 4 4,clip]{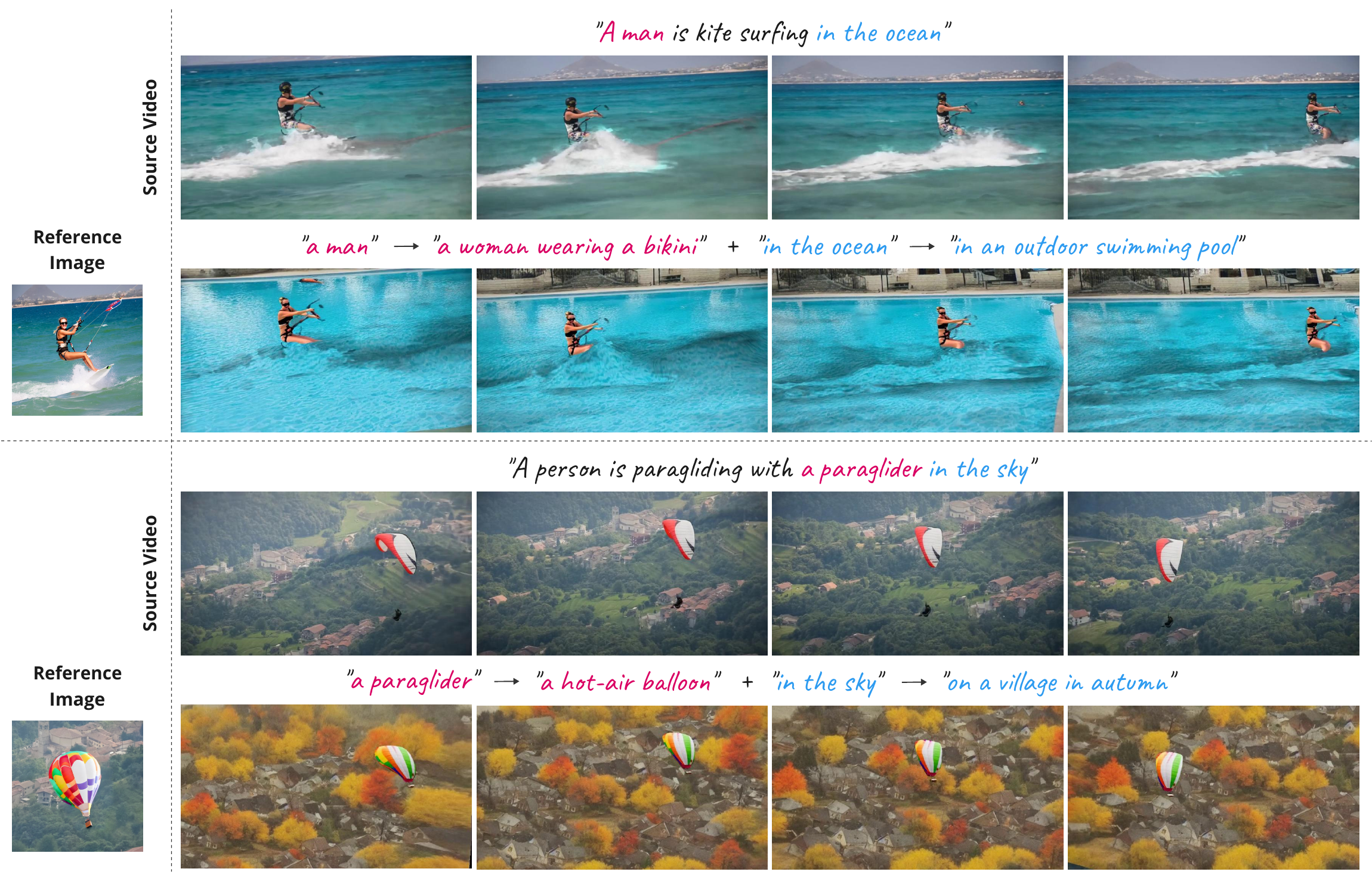}
    \caption{\textbf{Additional results of our method.}}
    \label{fig:more_results}
\end{figure*}

Besides these three losses, we leverage a pretrained diffusion model to provide pixel-level guidance by backpropagating the gradient of noise residual through the rendering process for fidelity improvement, inspired by DreamFusion~\cite{poole2022dreamfusion}. For each iteration, we randomly diffuse an edited frame $I^t_k$ with a noise $\varepsilon$ as the input for the pretrained diffusion model to output a predicted noise $\hat{\varepsilon}$. The gradient of noise residual $\hat{\varepsilon} - \varepsilon$ is then backpropagated to update the parameter set $\theta$ of $\sM_{A_{f}^{t}}$ and $\sM_{{\alpha}^t}$:
\begin{equation} \label{eq:sds}
    \nabla_{\theta}\gL_\mathrm{SDS}(I^t_k) = \mathbb{E}_{i, \varepsilon}[w(i)(\hat{\varepsilon}-\varepsilon) \cdot \nabla_{\theta}I^t_k] \,,
\end{equation}
where $w(i)$ is a weighting function depending on the timestep $i$. This process is called \textit{Score Distillation Sampling (SDS)}~\cite{poole2022dreamfusion}.
%
%However, these three losses are certainly not enough. We still need a constraint to guide how the object structures should be formed. Obviously, those edited RGB pixel values that are not located in the original foreground UV mapping areas can be mapped to totally different locations according to time $k$, as illustrated in \cref{fig:offset_loss}. To avoid non-foreground points being mapped unnaturally back to frames, causing unnecessary distortion, we encourage the offsets $\Delta$ between point $p(x, y, k)$ and $p'(x+\Delta, y+\Delta, k)$ to match the corresponding offsets on the optimized UV space according to the original UV mapping structure:
%
Finally, the total loss function is
\begin{equation} \label{eq:total_loss}
    \gL = \gL_\mathrm{SDS} + \lambda_\mathrm{rgb}\gL_\mathrm{rgb} + \lambda_{\alpha}\gL_{\alpha} + \lambda_\mathrm{off}\gL_\mathrm{off} \,.
\end{equation}

%% file: sec/4_experiments.tex
\section{Experiments}
\label{sec:experiments_4}
\subsection{Experimental setup}
We implement our diffusion atlas generating method over text-to-image (T2I) latent diffusion models~\cite{Rombach_2022_CVPR} with image embedding~\cite{Radford2021LearningTV} (Stable Diffusion UnCLIP v2-1\footnote{\url{https://huggingface.co/stabilityai/stable-diffusion-2-1-unclip}}). We apply our method on several videos from DAVIS dataset~\cite{Pont-Tuset_arXiv_2017}, with each video of 50 to 70 frames containing a moving foreground object. We center-crop and resize the object in atlas textures to the resolution of 768 $ \times$ 768 before feeding it into our model. We fine-tune the model for 150 steps on a learning rate $10^{-5}$. During inference, we apply DDIM sampler~\cite{song2020denoising} with classifier-free guidance~\cite{2022arXiv220712598H} with 50 steps. It takes about 10 minutes for fine-tuning and one minute for sampling on an A6000 GPU using 22GB of memory.

For the UV space optimization process, the resolution of each frame is set to 768 $\times$ 432. The MLPs $\mathbb{M}_{A_{f}^{t}}$ and $\mathbb{M}_{{\alpha}^t}$ exploit the architecture of LNA~\cite{kasten2021layered}. We use a batch size of 10,000 point coordinates and train each video for around 6,000 iterations. The optimization process takes about 40 minutes on an A6000 GPU using 16GB of memory.

\begin{figure*}[htp!]
    \centering
    \includegraphics[width=\textwidth, trim=4 4 4 4,clip]{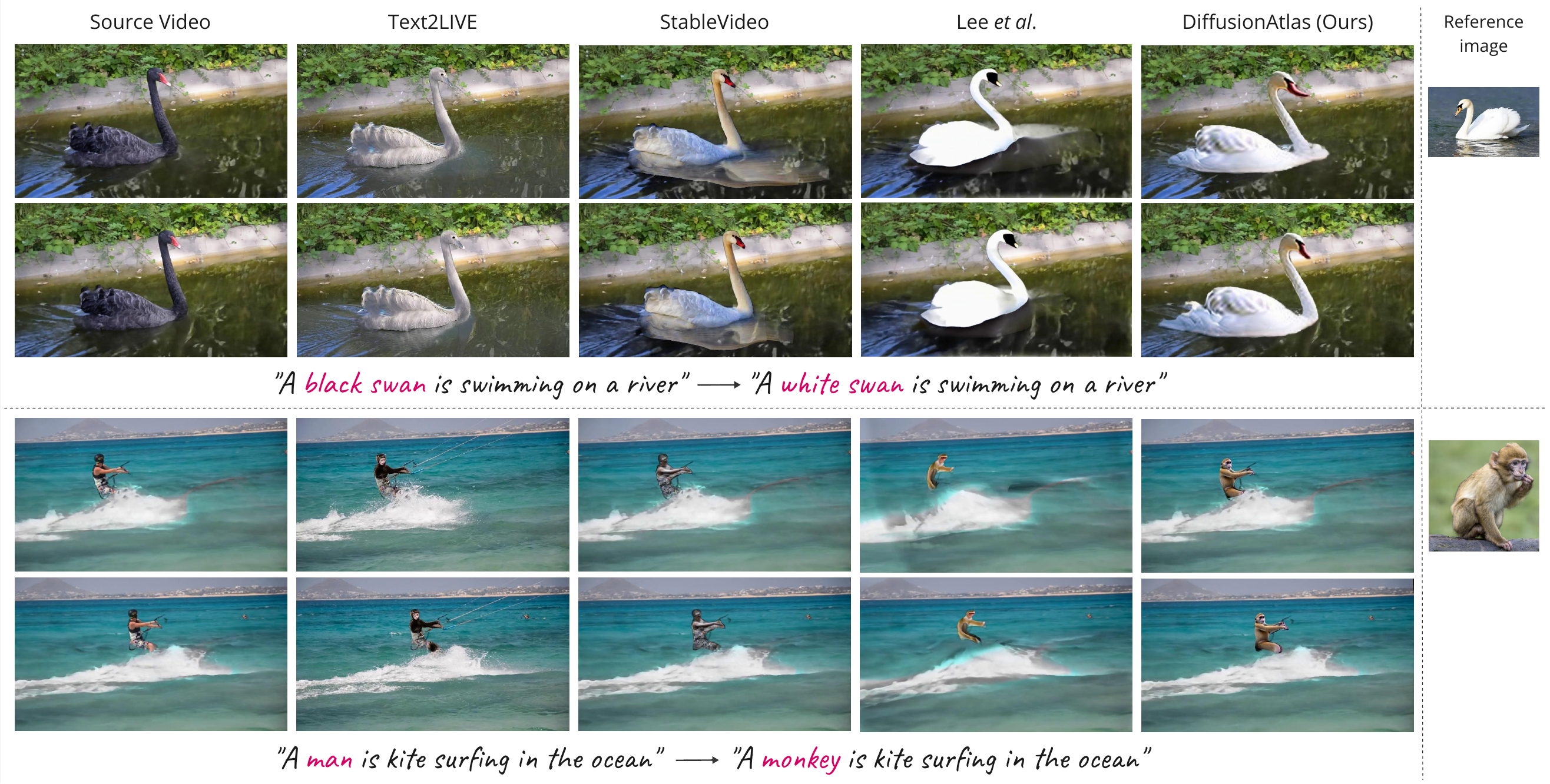}
    \caption{\textbf{Comparison with SOTA.} Our method outperforms other baselines in both object identity preservation and textual alignment.}
    \label{fig:sota_compare}
\end{figure*}

\subsection{Comparison with prior work}
\textbf{Compared SOTAs.} We compare our methods with three state-of-the-art atlas-based methods: 1) Text2LIVE~\cite{bar2022text2live} supports text-driven editing with a structure loss to preserve the original object shape. 2) StableVideo~\cite{chai2023stablevideo} develops an inter-frame propagation mechanism for object appearance consistency. 3) Lee \etal~\cite{lee2023textvideoedit} propose an approach to shape-aware video editing that employs semantic correspondence to transform the object shapes.

\textbf{Qualitative results.} \cref{fig:sota_compare} illustrates visual comparisons of our approach with SOTA atlas-based methods. Text2LIVE shows incomplete editing. This may be due to its reliance on creating a new editing layer, which leads to potential textual destruction in the entire video. While StableVideo achieves more holistic editing results than Text2LIVE, it still struggles to accurately generate objects that align well with the desired text prompt (\eg, its result of ``kite-surfing monkey'' may not closely resemble an actual monkey). Lee~\etal can manipulate the texture and shape of the objects to better match the desired editing conditions. However, their approach suffers from shape distortion and texture detail loss throughout the propagation process (\eg, its ``white swan'' result loses texture details on the wings). Such issues are an inherent trade-off due to the deformation formulation based on the semantic correspondence between objects, which leads to unnatural and inaccurate transformations. In contrast, our method outperforms them in high-fidelity editing, achieving a faithful representation of the desired prompts. Moreover, none of the methods above can perform video editing for both textual and visual cues.

\begin{table}
\centering
\resizebox{\columnwidth}{!}{
\begin{tabular}{@{}c@{\,}|@{\,}c@{\,}|@{\,}c@{\,}c@{}}
 & Metric Eval. & \multicolumn{2}{c}{User Preference} \\
 & CLIP score $\uparrow$ & Quality $\uparrow$ & Textual Alignment $\uparrow$ \\
\hline
 Text2LIVE & \textbf{31.23} & 36.45 & 27.6 \\
 StableVideo & 29.77 & 39.55 & 32.27 \\
 Lee \etal & 30.02 & 22.37 & 23.42 \\
 \hline
 Ours & 30.22 & (\textbf{63.55}, \textbf{60.45}, \textbf{77.63}) & (\textbf{72.4}, \textbf{67.73}, \textbf{76.58})
\end{tabular}
}
\caption{\textbf{Quantitative evaluation.} The triplet means preferences for (ours vs. Text2LIVE, ours vs. StableVideo, ours vs. Lee \etal)
}
\label{tab:clip_score}
\end{table}

\textbf{Quantitative results.} We quantify our method against SOTA in \Cref{tab:clip_score}. To measure textual alignment, we compute the CLIP score between the prompt and all frames and report the average score of 4 edited videos. In addition, we conduct a user study to compare the methods in terms of output quality and textual alignment. We present two videos generated by our method and a SOTA method in random order and ask 48 users ``Which video is more natural and realistic?" for video quality and "Which video is better described by the text?" for textual alignment. DiffusionAtlas achieves a comparable CLIP score as StableVideo and Lee \etal's method. Note that Text2LIVE explicitly optimizes a generator on a CLIP-based loss, making the metric not reliable for assessing its generalization performance and editing quality, as shown in the qualitative results. Regarding the user study, we find an overwhelming preference for DiffusionAtlas in both video quality and textual alignment, implying the superiority of our method. 

\begin{figure}[t!]
    \centering
    \begin{subfigure}[t]{0.47\textwidth}
        \centering
         \includegraphics[width=\textwidth,trim=4 4 4 4,clip]{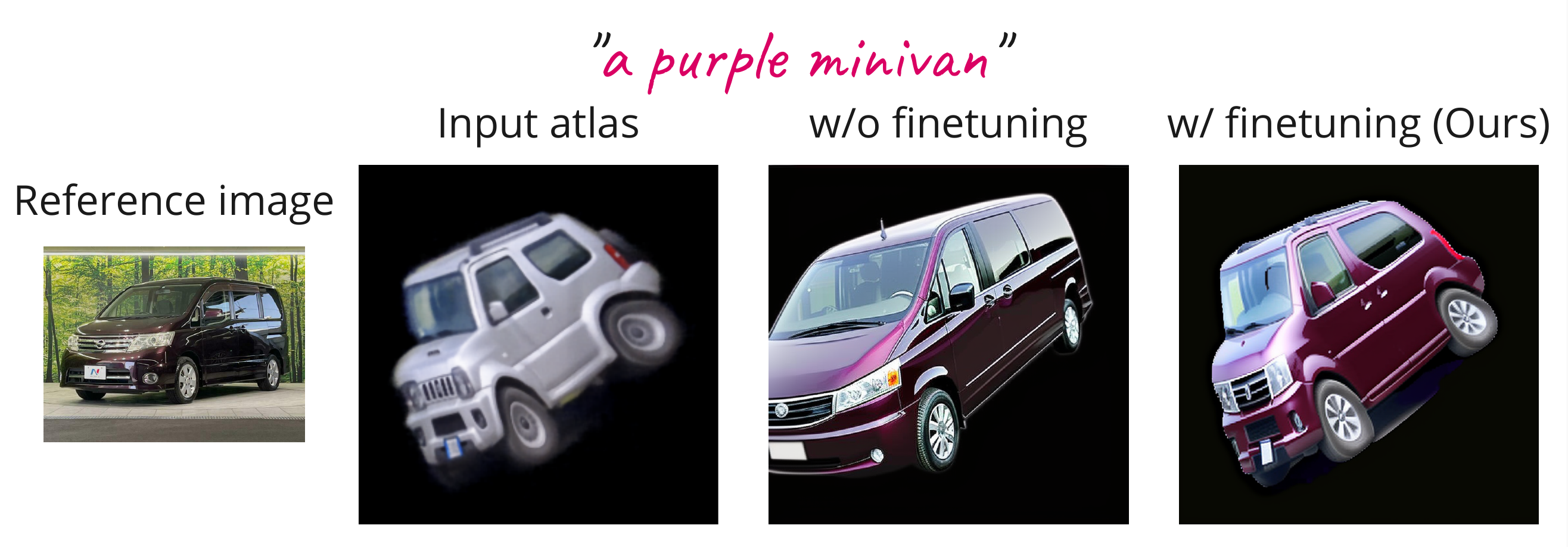}
         \caption{Editing atlas texture without the fine-tuning process results in a completely different object structure from the input atlas.}
         \label{fig:ablation_study(a)}
    \end{subfigure}
    \begin{subfigure}[t]{0.47\textwidth}
        \centering
         \includegraphics[width=\textwidth,trim=4 4 4 4,clip]{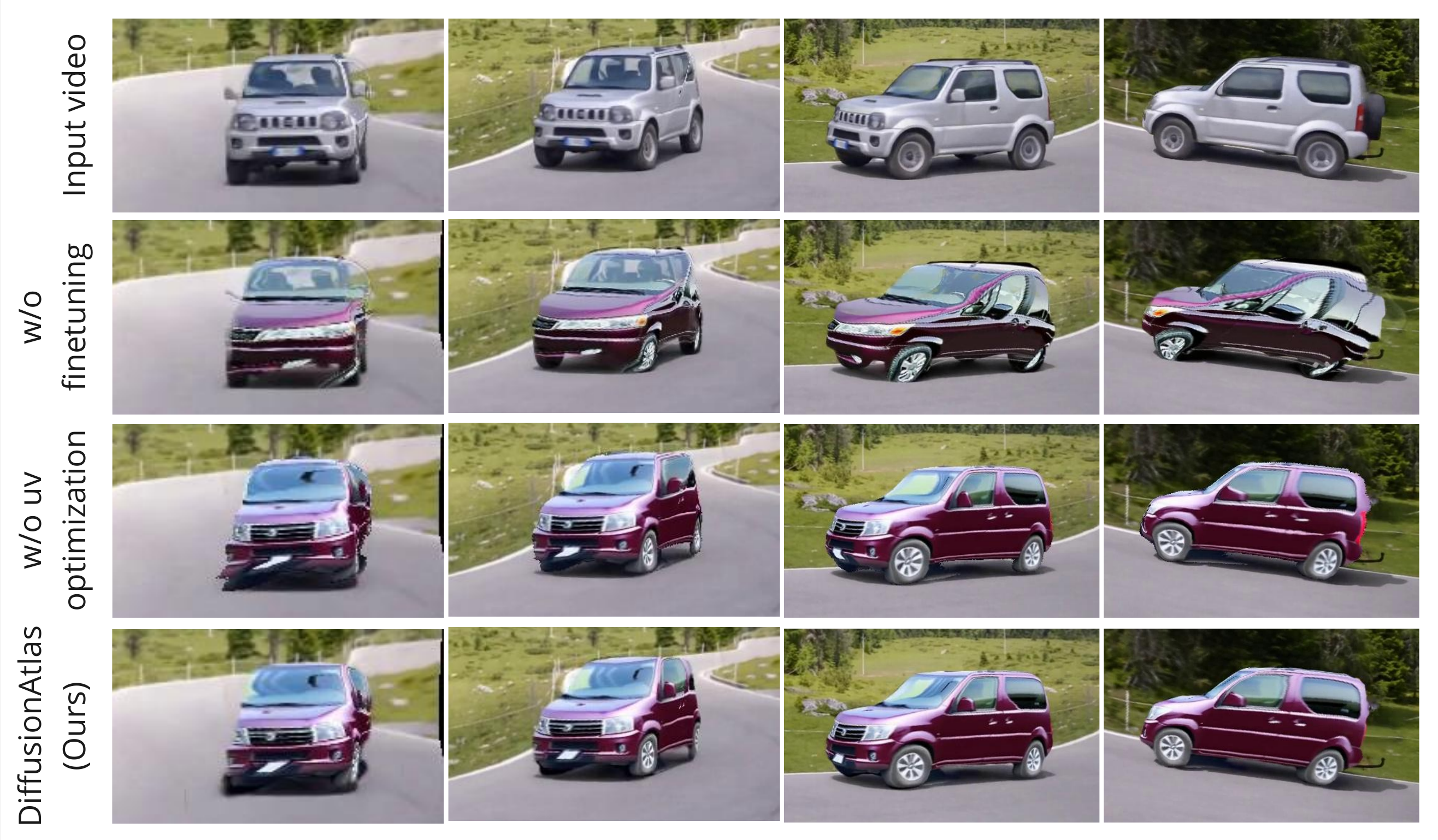}
         \caption{Without fine-tuning, the edited atlas yields severe distortions in the rendered video (the 2nd row). Without UV optimization, unnatural parts can still appear in the edited frames (the 3rd row). Our UV optimization process is crucial to guide and correct the distorted pixels (the last row).}
         \label{fig:ablation_study(b)}
    \end{subfigure}
    \caption{\textbf{Ablation study.} Ablations of removing the fine-tuning process (\cref{sec:method_3.2}) and the UV mapping optimization (\cref{sec:method_3.3}).}
    \label{fig:ablation_study}
    \vspace{-0.05in}
\end{figure}

\begin{figure}[t!]
    \centering
    \begin{subfigure}[t]{0.45\textwidth}
        \centering
         \includegraphics[width=\textwidth,trim=4 4 4 4,clip]{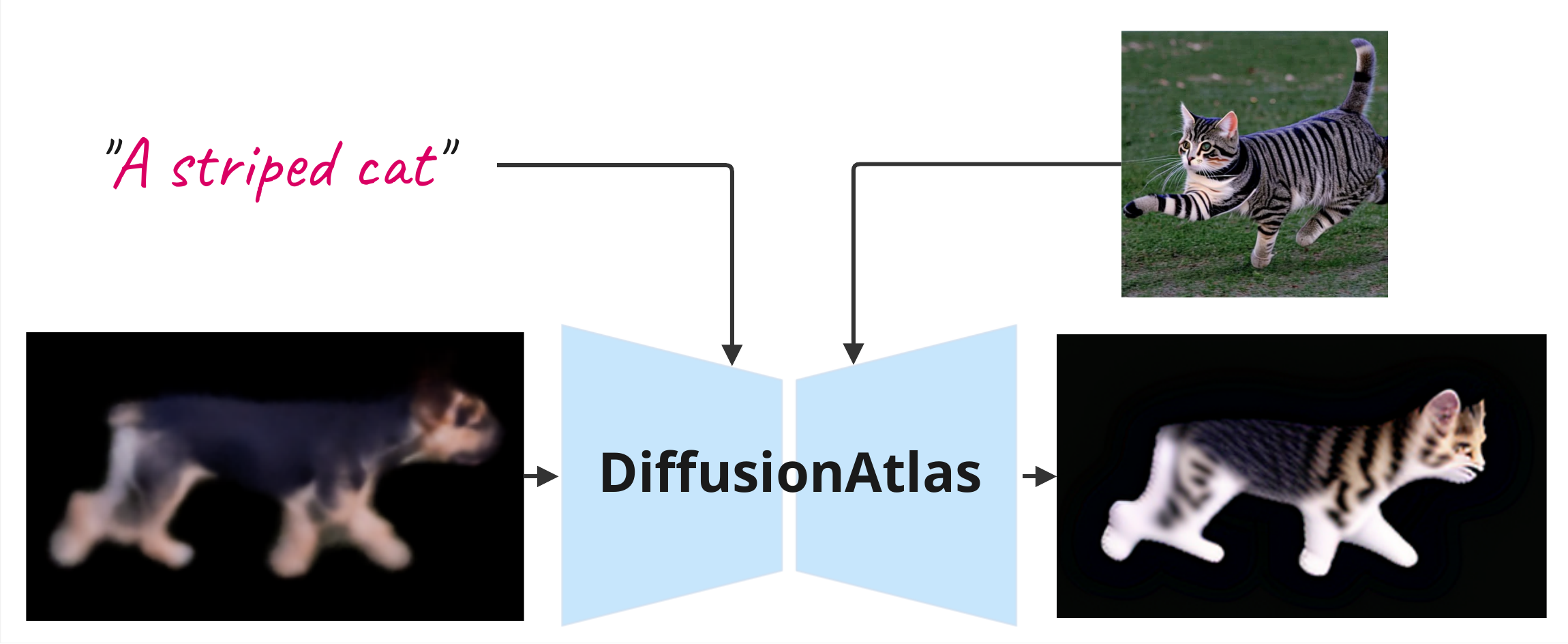}
         \caption{The fine-tuning process in DiffusionAtlas can cause overfitting to the source atlas, limiting huge changes in object shapes.}
         \label{fig:semantic_corr(a)}
    \end{subfigure}
    \begin{subfigure}[t]{0.45\textwidth}
        \centering
         \includegraphics[width=\textwidth,trim=4 4 4 4,clip]{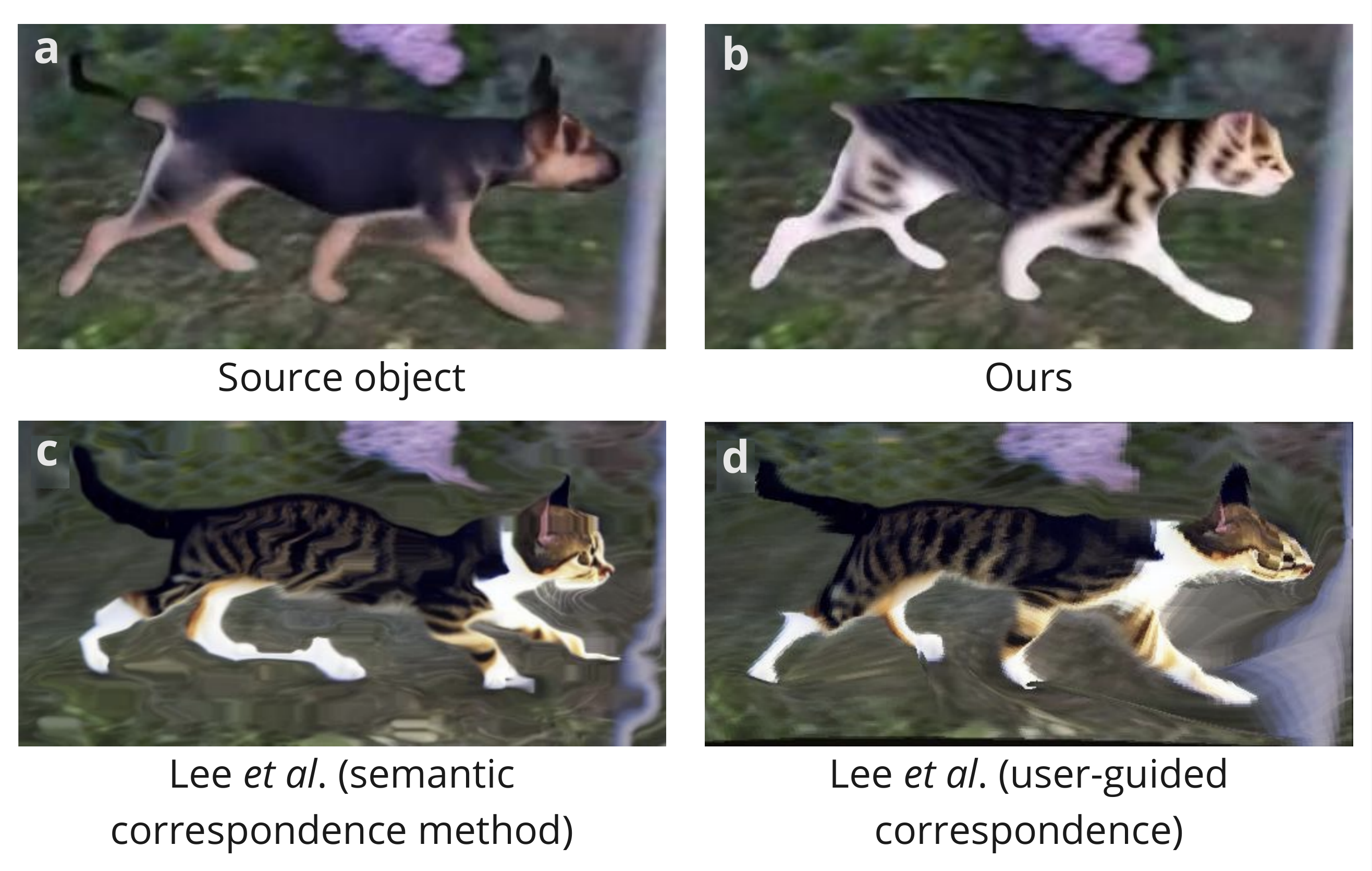}
         \caption{In editing ``a striped cat", our method resembles the semantic without feature matching~\cite{denseMatching} or manual correction as in Lee \etal~\cite{lee2023textvideoedit}.}
         \label{fig:semantic_corr(b)}
    \end{subfigure}
    \caption{\textbf{Limitations.} Our approach achieves high performance in semantic correspondence in an automatic way. Yet, it also limits our editing results in shape changes due to overfitting to the source atlas structure in the fine-tuning process.}
    \label{fig:semantic_corr}
    \vspace{-0.05in}
\end{figure}

\subsection{Ablation study}
\label{sec:ablation_study_4.3}
We conduct an ablation study in \cref{fig:ablation_study} to validate the importance of the fine-tuning process in our diffusion atlas editing from \cref{sec:method_3.2} and UV mapping optimization from \cref{sec:method_3.3}. \cref{fig:ablation_study(a)} shows that without fine-tuning, the model generates an edited atlas texture that does not match the structure of the input atlas, causing artifacts and distortions in the rendered frames, as shown in the second row in \cref{fig:ablation_study(b)}. However, propagating the edited atlas texture back to frames after fine-tuning via the original UV maps still yields artifacts in some areas, as shown in the third row in \cref{fig:ablation_study(b)}. Thus, by optimizing the UV maps, we refine the edited frames in a pixel-wise manner (e.g., the shape of the side-view mirror and wheels are corrected into more natural looks while the distortion of the car plate has been reduced).
% All loss function's ablation study in supplementary

\subsection{Application}
\label{sec:experiment_4.4}
Our method allows personalized editing with both textual and visual cues, as shown in \cref{fig:results_overview,fig:more_results}. For background editing, we directly apply text-driven image editing to the background atlas, treating it as a natural panorama image (see \cref{fig:model_pipeline}). To achieve high-fidelity in foreground object editing, we have fine-tuned our DiffusionAtlas model to generate target objects with similar structures. The fine-tuning establishes a robust semantic correspondence association between objects. In fact, correspondence emerges in image diffusion models, as the U-Net structure adeptly extracts features from the input image, as illustrated in~\cite{tang2023dift}. Our method excels in automatically dense matching between objects, resulting in more accurate object transformations without any additional semantic correspondence methods or manual corrections~\cite{lee2023textvideoedit} (see \cref{fig:semantic_corr(b)}).
% Metric for high performance in matching points (supplementary)

%% file: sec/5_limitations.tex
\section{Limitations}
\label{sec:limitation}
As described in \cref{sec:experiment_4.4}, our method can automatically match points between objects. Yet, this also limits the edited outcomes. First, our method is constrained by the capabilities of the pretrained text-to-image diffusion model. Second, our loss term from the UV optimization process exploits the original LNA mapping structures. These two aspects limit our results in large shape changes, as shown in \cref{fig:semantic_corr}: Given an input atlas of ``dog", a text prompt ``a striped cat", and a reference image, our method generates a striped cat atlas texture that semantically corresponds to the original dog atlas in a ``running pose". Yet, the cat is overfitted by the input atlas, causing the generated cat to have a texture similar to a mixture of the input dog and the reference gray-striped cat image. Moreover, the overfitting prevents any large shape changes.
% Future work?

%% file: sec/6_conclusion.tex
\section{Conclusion}
\label{sec:conclusion_6}
We have introduced DiffusionAtlas, an effective approach to high-fidelity personalized video editing. To preserve object identities regarding the input conditions, we propose a diffusion atlas editing procedure using a textual-and-visual-based diffusion model on the 2D layered video representations. We further refine the distorted regions by optimizing the UV mappings to propagate the edited atlas back to frames. Extensive experiments demonstrate superior results of our method compared with state-of-the-art approaches.

%% file: sec/X_supp_paper.tex
\clearpage
\twocolumn[{%
\renewcommand\twocolumn[1][]{#1}%
%\maketitle
\maketitlesupplementary
\begin{center}
    \centering
    \captionsetup{type=figure}
    \includegraphics[width=0.9\textwidth, trim=4 4 4 4,clip]{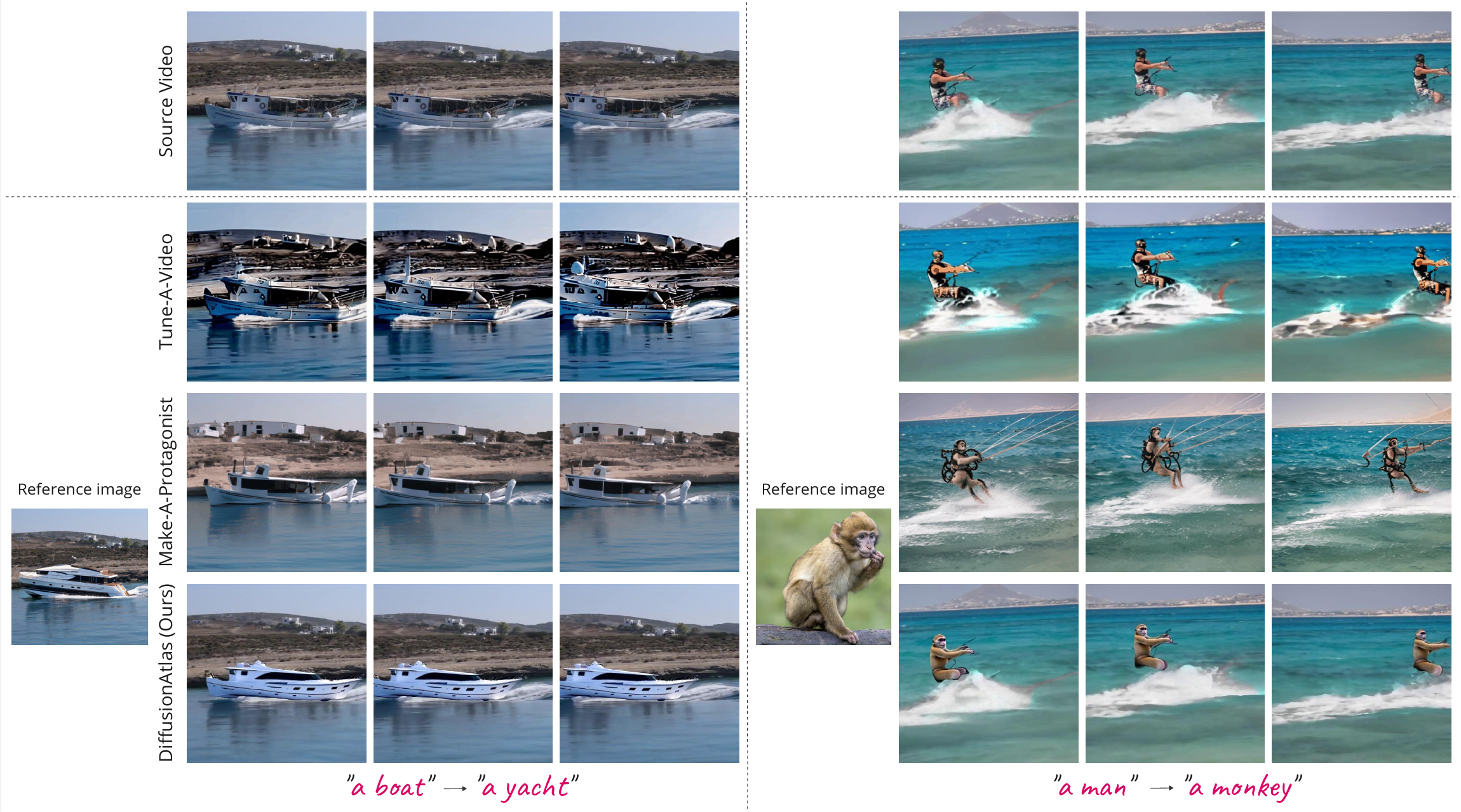}
    \captionof{figure}{\textbf{Additional qualitative results comparing our methods to Tune-A-Video and Make-A-Protagonist.} Note that we feed in the same reference image for both our method and Make-A-Protagonist.}
    \label{fig:additional_qualitative_results}
\end{center}%
}]

% User study detail
\section{Implementation detail}
\label{sec:imple_detail}
\subsection{DiffusionAtlas generation}
\label{sec:diffusionAtlas}

The original atlases obtained from the pretrained LNA~\cite{kasten2021layered} have a resolution of 1000 $ \times$ 1000. We center-crop the object in the foreground atlas and resize it to 768 $ \times$ 768 before feeding it into our DiffusionAtlas generation model. We set a maximum noise level for CLIP image embeddings to 50 to help the model memorize the input structure more efficiently during fine-tuning. Note that we full-tune the whole U-Net instead of only the attention blocks as in~\cite{wu2022tuneavideo, ruiz2023dreambooth}.
% Do I need to say why adding noise to image embedding?
% Do I need to show why full-tuning?

\subsection{UV mapping optimization}
\label{sec:UV_map_op}
In the UV mapping optimization process, all our networks are exploited from LNA~\cite{kasten2021layered}, which uses ReLU activation between hidden layers and a tanh function before outputting results. We use an Adam optimizer with a learning rate of $10^{-4}$ for optimizing all networks simultaneously. The atlas MLP $\sM_{A_{f}^{t}}$ has a short initial training phase, in which we train the identity mapping $(x,y,k) \rightarrow (x,y)$ for 100 iterations to retain the order of points.

In addition to our offset loss term $\gL_\mathrm{off}$, we implement a global offset loss in which we set the offset $\Delta$ to $5$ ($\Delta = 1$ for the original offset loss term). We train an additional 2000 iterations for the Score Distillation Sampling (SDS) $\gL_\mathrm{SDS}$ loss for final fidelity check.

Before propagating the atlas textures back to frames, we crop the objects on the edited foreground atlas texture $A^t_f$ back to the source foreground atlas $A^s_f$ to retain the moving shadows that are detected as foregrounds by the pre-trained LNA networks. For the undefined pixels that appear from shape changes, we apply a generative image inpainting technique, DeepFill~\cite{yu2018free}, to fill in the previously unseen pixels for smooth editing results.
% Add figures to demonstrate this.
% Weight of loss terms

\section{Additional comparison results}
\label{sec:add_results}
\begin{figure}[t!]
    \centering
    \begin{subfigure}[t]{0.47\textwidth}
        \centering
         \includegraphics[width=\textwidth,trim=4 4 4 4,clip]{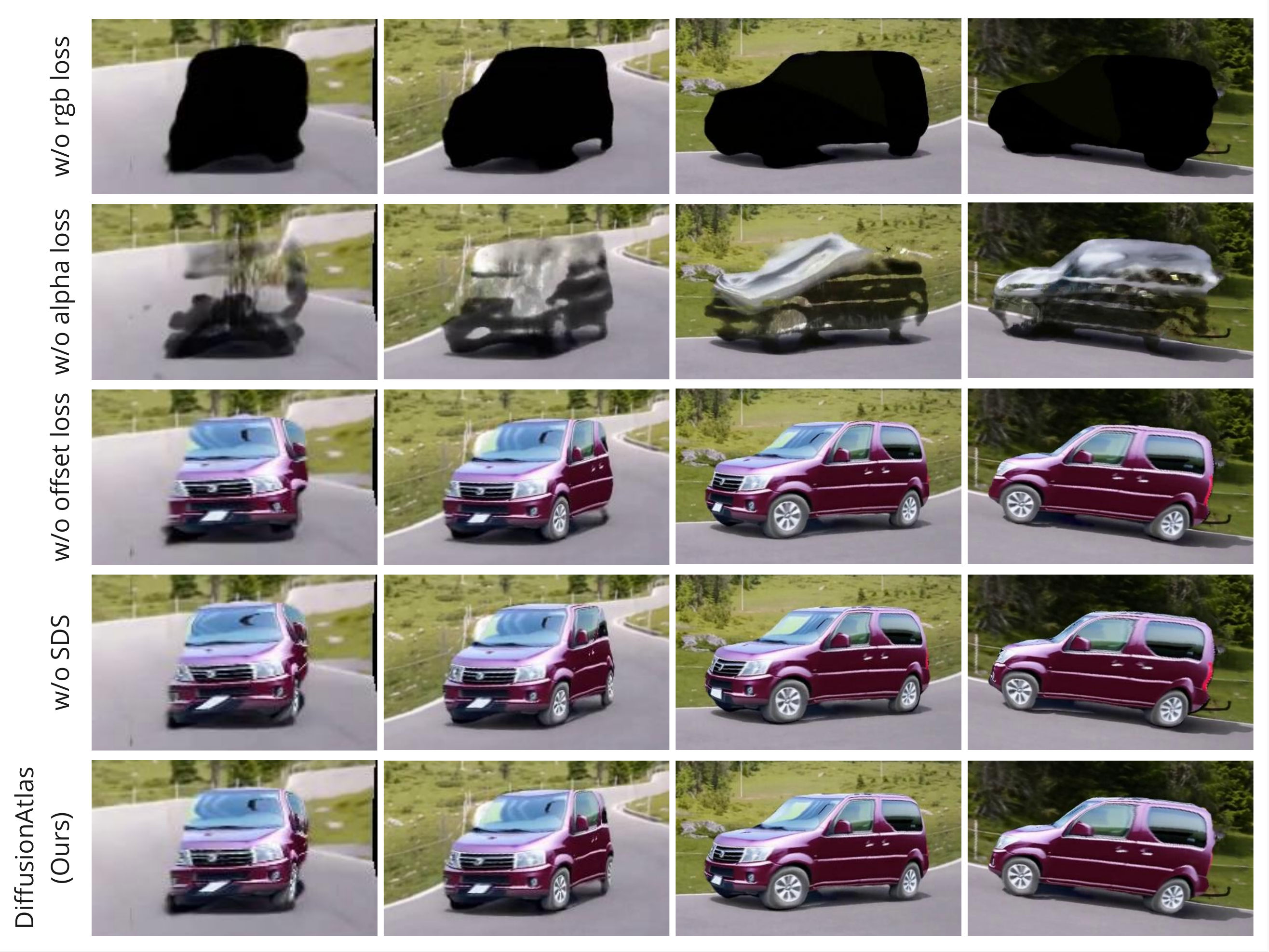}
         \caption{We show that all of our loss terms are crucial for the UV mapping optimization procedure.}
         \label{fig:uv_ablaition_study(a)}
    \end{subfigure}
    \begin{subfigure}[t]{0.47\textwidth}
        \centering
         \includegraphics[width=\textwidth,trim=4 4 4 4,clip]{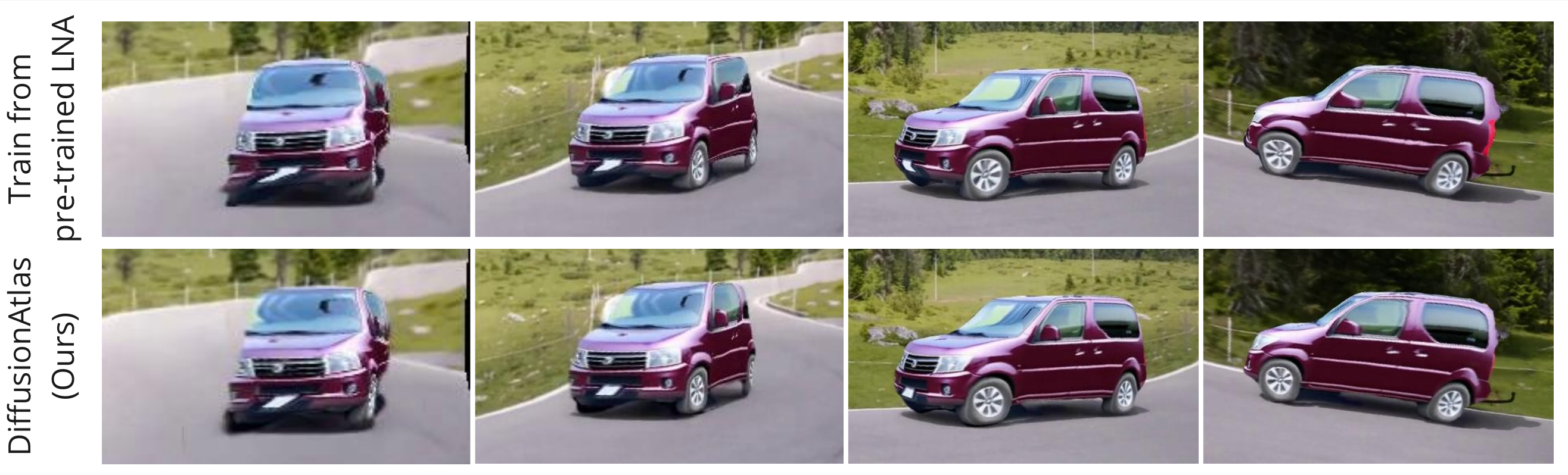}
         \caption{Fine-tuning from the pre-trained LNA networks does not lead to more promising editing results.}
         \label{fig:uv_ablaition_study(b)}
    \end{subfigure}
    \caption{\textbf{Additional ablation study on the loss terms of the UV mapping optimization process.}}
    \label{fig:uv_ablaition_study}
    \vspace{-0.05in}
\end{figure}

In this section, we compare our methods with two diffusion-based state-of-the-art methods: Tune-A-Video~\cite{wu2022tuneavideo} and Make-A-Protagonist~\cite{zhao2023makeaprotagonist}. Tune-A-Video~\cite{wu2022tuneavideo} is the first method to introduce the one-shot fine-tuning mechanism for T2V generation, while Make-A-Protagonist~\cite{zhao2023makeaprotagonist} allows both visual and textual cues for personalized generation.

\subsection{Qualitative results}
\label{sec:add_quati_results}
We present a visual comparison of our approach with the two aforementioned methods in \cref{fig:additional_qualitative_results}. Both Tune-A-Video and Make-A-Protagonist are not able to preserve non-target areas with flickering artifacts appearing in their generated videos (\eg, the background of the example "yacht"). Our method is able to achieve better spatiotemporal consistency since we generate new objects on the 2D layered representations of the objects. Additionally, both methods fail to attain high fidelity on the target object identities (\eg, both approaches fail to properly generate "monkey" based on either the text prompt or the reference image).

\subsection{Quantitative results}
\label{sec:add_quanti_results}
\begin{table}
\centering
\resizebox{\columnwidth}{!}{
\begin{tabular}{c|c|c}
 & Textual Alignment $\uparrow$ & Object Fidelity $\uparrow$ \\
\hline
 Tune-A-Video & 29.48 & - \\
 Make-A-Protagonist & 29.03 & 71.71 \\
 \hline
 DiffusionAtlas (Ours) & \textbf{29.57} & \textbf{76.09}
\end{tabular}
}
\caption{\textbf{Quantitative evaluation.}}
\label{tab:clip_score}
\end{table}

\begin{figure}[t!]
    \centering
    \includegraphics[width=0.47\textwidth,trim=4 4 4 4,clip]{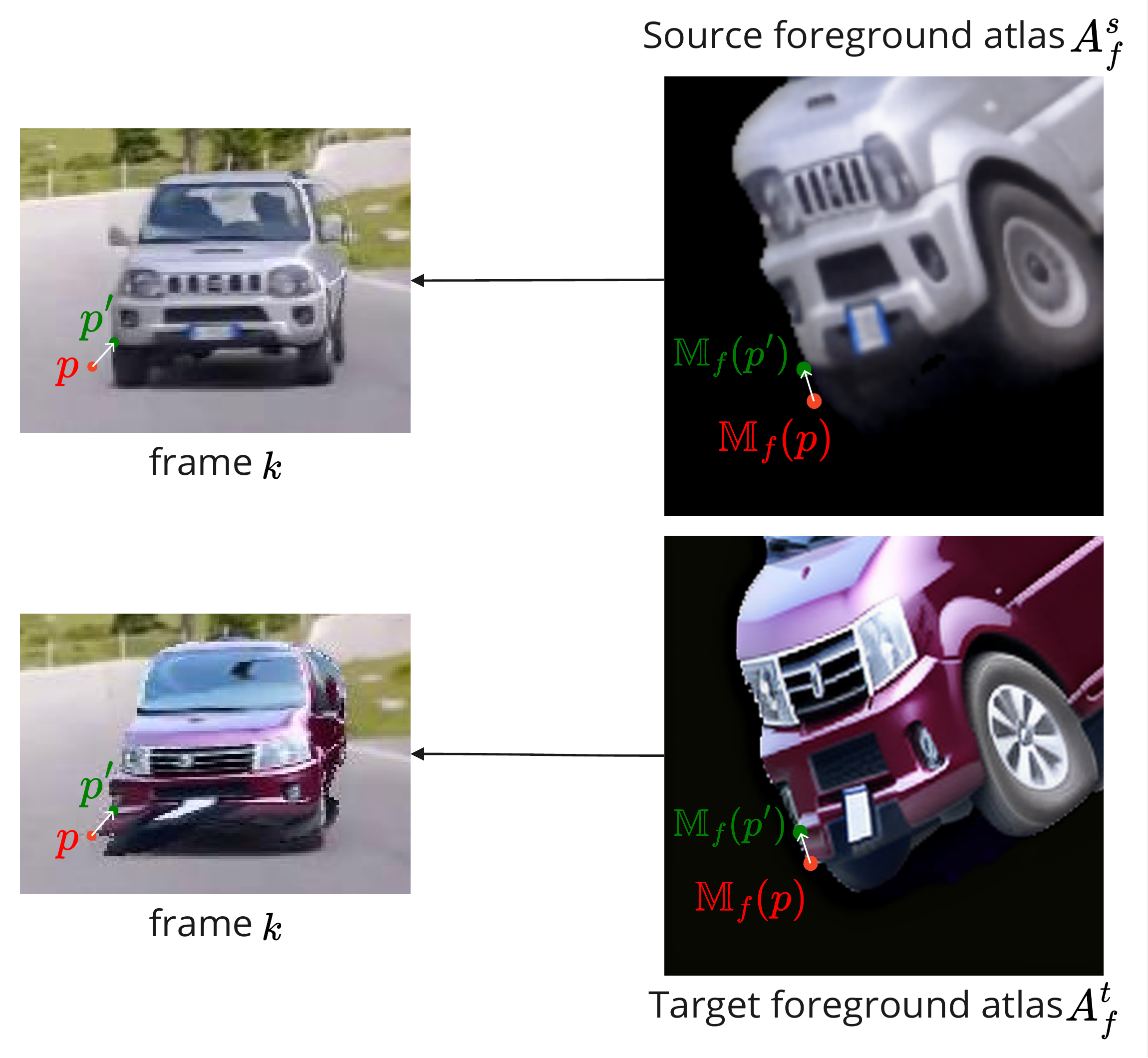}
    \caption{\textbf{Illustration of the offset loss.} A point $p$ in frame $k$ that is within the non-foreground area in the source UV space and its offset point $p'$ can lead to texture shifting with the generated target atlas texture. Therefore, we apply the offset loss on the atlas space for more flexible optimization. (Note that the atlas textures shown here are cropped and zoomed in for clearer visualization.)}
    \label{fig:offset_loss_supp}
\end{figure}

\begin{figure*}[htp!]
    \centering
    \includegraphics[width=\textwidth, trim=4 4 4 4,clip]{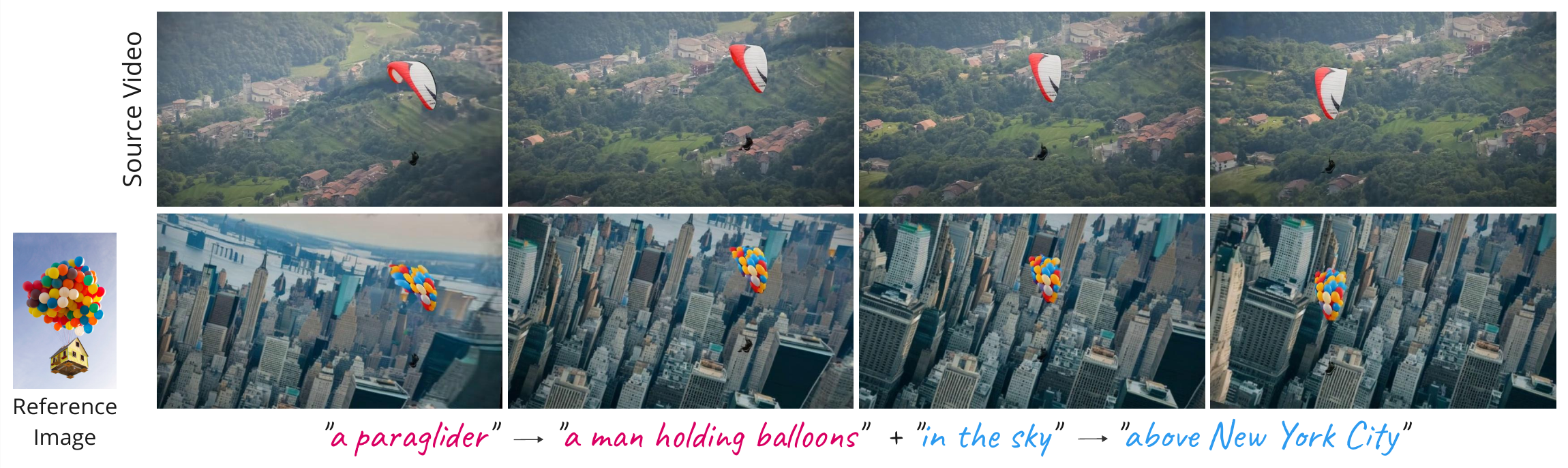}
    \caption{\textbf{Additional editing results.}}
    \label{fig:additional_results}
\end{figure*}

\Cref{tab:clip_score} shows our quantitative results compared to the two methods. To measure textual alignment, we compute the average CLIP score between all frames of the two example videos in \cref{fig:additional_qualitative_results} and corresponding edited prompts. For object fidelity, we calculate the average CLIP score between frames and the reference image embedding. We demonstrate that our method outperforms in both textual alignment and object fidelity compared to Tune-A-Video and Make-A-Protagonist.

\section{Additional ablation study on loss terms}
\label{sec:more_abla_study}

We conduct an ablation study on the loss terms of the UV mapping optimization process in \cref{fig:uv_ablaition_study(a)}. It is obvious that without the two reconstruction losses, RGB loss $\gL_\mathrm{rgb}$ and alpha loss $\gL_{\alpha}$, the object cannot even be formed on the frames as shown in the first and second row of \cref{fig:uv_ablaition_study(a)}. The Score Distillation Sampling (SDS) is crucial for our optimization process since it is capable of correcting shape distortions (\eg, see the plate, the wheels, and the windows in the fourth row in \cref{fig:uv_ablaition_study(a)}).

The main core behind the offset loss $\gL_\mathrm{off}$ is to connect pixel locations in frames to the coordinates of the optimized UV space in a more structural and logical way. We consider a point $p$ in frame $k$ that is a point in the non-foreground area, meaning that the corresponding point of $p$ in atlas space, say $\sM_{f}(p)$ with the pre-trained LNA~\cite{kasten2021layered}, does not belong to the source foreground UV space. This kind of point will lead to shape distortion and texture deviation, as illustrated in \cref{fig:offset_loss_supp}. In order to guide the networks to connect points between non-foreground and foreground areas more naturally, we compute the loss term in the atlas space, enabling offsets between pixel locations in frames to form in the correct direction and structure (\eg, $\sM_{f}(p)$ and $\sM_{f}(p')$ in \cref{fig:offset_loss_supp}).

In \cref{fig:uv_ablaition_study(b)}, we show that fine-tuning from the original pre-trained LNA~\cite{kasten2021layered} does not result in more promising edited outputs. This might be caused by the weight being heavily fixed by the original correlation between frames and the UV space, causing constraints in modifying and correcting shape distortions and texture shiftings when editing a new object. Consequently, we show that our offset loss is important for the optimization process, as it can help the model connect frame pixel locations to UV space coordinates without being constrained by the original correlation. As shown in the third row in \cref{fig:uv_ablaition_study(a)}, without the offset loss $\gL_\mathrm{off}$, the networks do not know how to form the object structure in some areas (\eg, missing parts appear from the minivan when it is making a turn).

\section{Additional results}
\label{sec:additional_results}
\begin{figure*}[htp!]
    \centering
    \includegraphics[width=\textwidth, trim=4 4 4 4,clip]{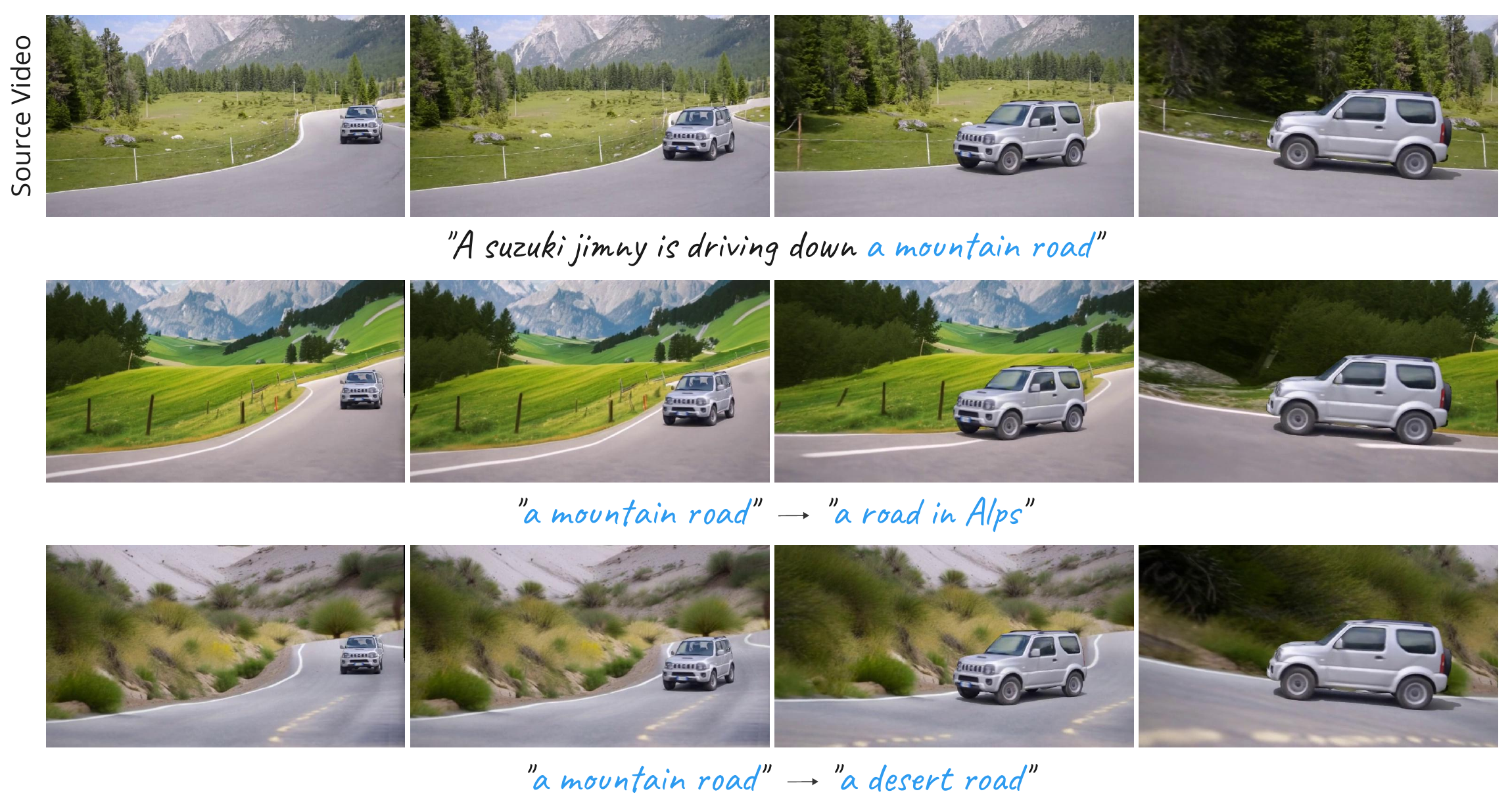}
    \caption{\textbf{Additional results on background editing.}}
    \label{fig:bg_edit_results}
\end{figure*}

We provide more editing results to demonstrate the effectiveness of our approach. \cref{fig:additional_results} showcases additional video editing results of our method, and \cref{fig:bg_edit_results} shows additional results on background editing.

\section{Future work}
\label{sec:future_work}
As outlined in Sec.~\ref{sec:limitation} of the main paper, the main limitation of our method is that it tends to overfit the structure of the source atlas in the fine-tuning phase. A viable solution is to incorporate spatial transformation between objects in the DiffusionAtlas generation process, enabling diffusion models to obtain both the semantic correspondence information and be aware of the shape changes when generating new object atlases. The exploration of this research direction is left to future work.

%% file: main.bbl
\begin{thebibliography}{55}
\providecommand{\natexlab}[1]{#1}
\providecommand{\url}[1]{\texttt{#1}}
\expandafter\ifx\csname urlstyle\endcsname\relax
  \providecommand{\doi}[1]{doi: #1}\else
  \providecommand{\doi}{doi: \begingroup \urlstyle{rm}\Url}\fi

\bibitem[Bar-Tal et~al.(2022)Bar-Tal, Ofri-Amar, Fridman, Kasten, and Dekel]{bar2022text2live}
Omer Bar-Tal, Dolev Ofri-Amar, Rafail Fridman, Yoni Kasten, and Tali Dekel.
\newblock Text2live: Text-driven layered image and video editing.
\newblock In \emph{European Conference on Computer Vision}, pages 707--723. Springer, 2022.

\bibitem[Brooks et~al.(2023)Brooks, Holynski, and Efros]{brooks2022instructpix2pix}
Tim Brooks, Aleksander Holynski, and Alexei~A. Efros.
\newblock Instructpix2pix: Learning to follow image editing instructions.
\newblock In \emph{CVPR}, 2023.

\bibitem[Chai et~al.(2023)Chai, Guo, Wang, and Lu]{chai2023stablevideo}
Wenhao Chai, Xun Guo, Gaoang Wang, and Yan Lu.
\newblock Stablevideo: Text-driven consistency-aware diffusion video editing.
\newblock \emph{arXiv preprint arXiv:2308.09592}, 2023.

\bibitem[Couairon et~al.(2023)Couairon, Rambour, Haugeard, and Thome]{couairon2023videdit}
Paul Couairon, Cl{\'{e}}ment Rambour, Jean{-}Emmanuel Haugeard, and Nicolas Thome.
\newblock Videdit: Zero-shot and spatially aware text-driven video editing.
\newblock \emph{CoRR}, abs/2306.08707, 2023.

\bibitem[Dhariwal and Nichol(2021)]{NEURIPS2021_49ad23d1}
Prafulla Dhariwal and Alexander Nichol.
\newblock Diffusion models beat gans on image synthesis.
\newblock In \emph{Advances in Neural Information Processing Systems}, pages 8780--8794. Curran Associates, Inc., 2021.

\bibitem[et~al(2022)]{ramesh2022}
Aditya~Ramesh et al.
\newblock Hierarchical text-conditional image generation with clip latents, 2022.

\bibitem[Gal et~al.(2021)Gal, Patashnik, Maron, Chechik, and Cohen-Or]{gal2021stylegannada}
Rinon Gal, Or Patashnik, Haggai Maron, Gal Chechik, and Daniel Cohen-Or.
\newblock Stylegan-nada: Clip-guided domain adaptation of image generators, 2021.

\bibitem[Gal et~al.(2022)Gal, Alaluf, Atzmon, Patashnik, Bermano, Chechik, and Cohen-Or]{gal2022textual}
Rinon Gal, Yuval Alaluf, Yuval Atzmon, Or Patashnik, Amit~H. Bermano, Gal Chechik, and Daniel Cohen-Or.
\newblock An image is worth one word: Personalizing text-to-image generation using textual inversion, 2022.

\bibitem[Hertz et~al.(2023)Hertz, Mokady, Tenenbaum, Aberman, Pritch, and Cohen{-}Or]{hertz2022prompt}
Amir Hertz, Ron Mokady, Jay Tenenbaum, Kfir Aberman, Yael Pritch, and Daniel Cohen{-}Or.
\newblock Prompt-to-prompt image editing with cross-attention control.
\newblock In \emph{ICLR}. OpenReview.net, 2023.

\bibitem[{Ho} and {Salimans}(2022)]{2022arXiv220712598H}
Jonathan {Ho} and Tim {Salimans}.
\newblock {Classifier-Free Diffusion Guidance}.
\newblock \emph{arXiv e-prints}, art. arXiv:2207.12598, 2022.

\bibitem[Ho et~al.(2020)Ho, Jain, and Abbeel]{ho2020denoising}
Jonathan Ho, Ajay Jain, and Pieter Abbeel.
\newblock Denoising diffusion probabilistic models.
\newblock \emph{arXiv preprint arxiv:2006.11239}, 2020.

\bibitem[Ho et~al.(2022{\natexlab{a}})Ho, Chan, Saharia, Whang, Gao, Gritsenko, Kingma, Poole, Norouzi, Fleet, and Salimans]{imagen_video}
Jonathan Ho, William Chan, Chitwan Saharia, Jay Whang, Ruiqi Gao, Alexey~A. Gritsenko, Diederik~P. Kingma, Ben Poole, Mohammad Norouzi, David~J. Fleet, and Tim Salimans.
\newblock Imagen video: High definition video generation with diffusion models.
\newblock \emph{CoRR}, abs/2210.02303, 2022{\natexlab{a}}.

\bibitem[Ho et~al.(2022{\natexlab{b}})Ho, Salimans, Gritsenko, Chan, Norouzi, and Fleet]{ho2022video}
Jonathan Ho, Tim Salimans, Alexey Gritsenko, William Chan, Mohammad Norouzi, and David~J Fleet.
\newblock Video diffusion models.
\newblock \emph{arXiv:2204.03458}, 2022{\natexlab{b}}.

\bibitem[Hu et~al.(2022)Hu, Shen, Wallis, Allen-Zhu, Li, Wang, Wang, and Chen]{hu2022lora}
Edward~J Hu, Yelong Shen, Phillip Wallis, Zeyuan Allen-Zhu, Yuanzhi Li, Shean Wang, Lu Wang, and Weizhu Chen.
\newblock Lo{RA}: Low-rank adaptation of large language models.
\newblock In \emph{International Conference on Learning Representations}, 2022.

\bibitem[Jamri\v{s}ka et~al.(2019)Jamri\v{s}ka, \v{S}\'{a}rka Sochorov\'{a}, Texler, Luk\'{a}\v{c}, Fi\v{s}er, Lu, Shechtman, and S\'{y}kora]{Jamriska19-SIG}
Ond\v{r}ej Jamri\v{s}ka, \v{S}\'{a}rka Sochorov\'{a}, Ond\v{r}ej Texler, Michal Luk\'{a}\v{c}, Jakub Fi\v{s}er, Jingwan Lu, Eli Shechtman, and Daniel S\'{y}kora.
\newblock Stylizing video by example.
\newblock \emph{ACM Transactions on Graphics}, 38\penalty0 (4), 2019.

\bibitem[Kasten et~al.(2021)Kasten, Ofri, Wang, and Dekel]{kasten2021layered}
Yoni Kasten, Dolev Ofri, Oliver Wang, and Tali Dekel.
\newblock Layered neural atlases for consistent video editing.
\newblock \emph{ACM Transactions on Graphics (TOG)}, 40\penalty0 (6):\penalty0 1--12, 2021.

\bibitem[Kawar et~al.(2023)Kawar, Zada, Lang, Tov, Chang, Dekel, Mosseri, and Irani]{kawar2023imagic}
Bahjat Kawar, Shiran Zada, Oran Lang, Omer Tov, Huiwen Chang, Tali Dekel, Inbar Mosseri, and Michal Irani.
\newblock Imagic: Text-based real image editing with diffusion models.
\newblock In \emph{Conference on Computer Vision and Pattern Recognition 2023}, 2023.

\bibitem[Kim et~al.(2022)Kim, Kwon, and Ye]{Kim_2022_CVPR}
Gwanghyun Kim, Taesung Kwon, and Jong~Chul Ye.
\newblock Diffusionclip: Text-guided diffusion models for robust image manipulation.
\newblock In \emph{Proceedings of the IEEE/CVF Conference on Computer Vision and Pattern Recognition (CVPR)}, pages 2426--2435, 2022.

\bibitem[Kirillov et~al.(2023)Kirillov, Mintun, Ravi, Mao, Rolland, Gustafson, Xiao, Whitehead, Berg, Lo, Doll{\'a}r, and Girshick]{kirillov2023segany}
Alexander Kirillov, Eric Mintun, Nikhila Ravi, Hanzi Mao, Chloe Rolland, Laura Gustafson, Tete Xiao, Spencer Whitehead, Alexander~C. Berg, Wan-Yen Lo, Piotr Doll{\'a}r, and Ross Girshick.
\newblock Segment anything.
\newblock \emph{arXiv:2304.02643}, 2023.

\bibitem[Lee et~al.(2023)Lee, Jang, Chen, Qiu, and Huang]{lee2023textvideoedit}
Yao-Chih Lee, Ji-Ze Genevieve~Jang Jang, Yi-Ting Chen, Elizabeth Qiu, and Jia-Bin Huang.
\newblock Shape-aware text-driven layered video editing demo.
\newblock \emph{arXiv preprint arXiv:2301.13173}, 2023.

\bibitem[Li et~al.(2020)Li, Qi, Lukasiewicz, and Torr]{li2020manigan}
Bowen Li, Xiaojuan Qi, Thomas Lukasiewicz, and Philip~HS Torr.
\newblock Manigan: Text-guided image manipulation.
\newblock In \emph{Proceedings of the IEEE/CVF Conference on Computer Vision and Pattern Recognition}, pages 7880--7889, 2020.

\bibitem[Li et~al.(2019)Li, Zhang, Zhang, Huang, He, Lyu, and Gao]{objgan19}
Wenbo Li, Pengchuan Zhang, Lei Zhang, Qiuyuan Huang, Xiaodong He, Siwei Lyu, and Jianfeng Gao.
\newblock Object-driven text-to-image synthesis via adversarial training.
\newblock In \emph{CVPR}, pages 12174--12182. Computer Vision Foundation / {IEEE}, 2019.

\bibitem[Liao et~al.(2021)Liao, Hu, Yang, and Rosenhahn]{liao2021text}
Wentong Liao, Kai Hu, Michael~Ying Yang, and Bodo Rosenhahn.
\newblock Text to image generation with semantic-spatial aware gan.
\newblock \emph{arXiv preprint arXiv:2104.00567}, 2021.

\bibitem[Liu et~al.(2023{\natexlab{a}})Liu, Zeng, Ren, Li, Zhang, Yang, Li, Yang, Su, Zhu, et~al.]{liu2023grounding}
Shilong Liu, Zhaoyang Zeng, Tianhe Ren, Feng Li, Hao Zhang, Jie Yang, Chunyuan Li, Jianwei Yang, Hang Su, Jun Zhu, et~al.
\newblock Grounding dino: Marrying dino with grounded pre-training for open-set object detection.
\newblock \emph{arXiv preprint arXiv:2303.05499}, 2023{\natexlab{a}}.

\bibitem[Liu et~al.(2023{\natexlab{b}})Liu, Zhang, Li, Lin, and Jia]{liu2023videop2p}
Shaoteng Liu, Yuechen Zhang, Wenbo Li, Zhe Lin, and Jiaya Jia.
\newblock Video-p2p: Video editing with cross-attention control.
\newblock \emph{arXiv:2303.04761}, 2023{\natexlab{b}}.

\bibitem[Lu et~al.(2020)Lu, Cole, Dekel, Xie, Zisserman, Salesin, Freeman, and Rubinstein]{lu2020}
Erika Lu, Forrester Cole, Tali Dekel, Weidi Xie, Andrew Zisserman, David Salesin, William~T Freeman, and Michael Rubinstein.
\newblock Layered neural rendering for retiming people in video.
\newblock In \emph{SIGGRAPH Asia}, 2020.

\bibitem[Lu et~al.(2021)Lu, Cole, Dekel, Zisserman, Freeman, and Rubinstein]{lu2021}
Erika Lu, Forrester Cole, Tali Dekel, Andrew Zisserman, William~T Freeman, and Michael Rubinstein.
\newblock Omnimatte: Associating objects and their effects in video.
\newblock In \emph{CVPR}, 2021.

\bibitem[Lu et~al.(2022)Lu, Cole, Dekel, Xie, Zisserman, Freeman, and Rubinstein]{lu2022}
Erika Lu, Forrester Cole, Tali Dekel, Weidi Xie, Andrew Zisserman, William~T Freeman, and Michael Rubinstein.
\newblock Associating objects and their effects in video through coordination games.
\newblock In \emph{NeurIPS}, 2022.

\bibitem[Meng et~al.(2022)Meng, He, Song, Song, Wu, Zhu, and Ermon]{meng2022sdedit}
Chenlin Meng, Yutong He, Yang Song, Jiaming Song, Jiajun Wu, Jun-Yan Zhu, and Stefano Ermon.
\newblock {SDE}dit: Guided image synthesis and editing with stochastic differential equations.
\newblock In \emph{International Conference on Learning Representations}, 2022.

\bibitem[Molad et~al.(2023)Molad, Horwitz, Valevski, Acha, Matias, Pritch, Leviathan, and Hoshen]{molad2023dreamix}
Eyal Molad, Eliahu Horwitz, Dani Valevski, Alex~Rav Acha, Yossi Matias, Yael Pritch, Yaniv Leviathan, and Yedid Hoshen.
\newblock Dreamix: Video diffusion models are general video editors.
\newblock \emph{arXiv preprint arXiv:2302.01329}, 2023.

\bibitem[Nichol et~al.(2022)Nichol, Dhariwal, Ramesh, Shyam, Mishkin, Mcgrew, Sutskever, and Chen]{pmlr-v162-nichol22a}
Alexander~Quinn Nichol, Prafulla Dhariwal, Aditya Ramesh, Pranav Shyam, Pamela Mishkin, Bob Mcgrew, Ilya Sutskever, and Mark Chen.
\newblock {GLIDE}: Towards photorealistic image generation and editing with text-guided diffusion models.
\newblock In \emph{Proceedings of the 39th International Conference on Machine Learning}, pages 16784--16804. PMLR, 2022.

\bibitem[Patashnik et~al.(2021)Patashnik, Wu, Shechtman, Cohen-Or, and Lischinski]{Patashnik_2021_ICCV}
Or Patashnik, Zongze Wu, Eli Shechtman, Daniel Cohen-Or, and Dani Lischinski.
\newblock Styleclip: Text-driven manipulation of stylegan imagery.
\newblock In \emph{Proceedings of the IEEE/CVF International Conference on Computer Vision (ICCV)}, pages 2085--2094, 2021.

\bibitem[Pont-Tuset et~al.(2017)Pont-Tuset, Perazzi, Caelles, Arbel\'aez, Sorkine-Hornung, and {Van Gool}]{Pont-Tuset_arXiv_2017}
Jordi Pont-Tuset, Federico Perazzi, Sergi Caelles, Pablo Arbel\'aez, Alexander Sorkine-Hornung, and Luc {Van Gool}.
\newblock The 2017 davis challenge on video object segmentation.
\newblock \emph{arXiv:1704.00675}, 2017.

\bibitem[Poole et~al.(2022)Poole, Jain, Barron, and Mildenhall]{poole2022dreamfusion}
Ben Poole, Ajay Jain, Jonathan~T. Barron, and Ben Mildenhall.
\newblock Dreamfusion: Text-to-3d using 2d diffusion.
\newblock \emph{arXiv}, 2022.

\bibitem[Qi et~al.(2023)Qi, Cun, Zhang, Lei, Wang, Shan, and Chen]{qi2023fatezero}
Chenyang Qi, Xiaodong Cun, Yong Zhang, Chenyang Lei, Xintao Wang, Ying Shan, and Qifeng Chen.
\newblock Fatezero: Fusing attentions for zero-shot text-based video editing.
\newblock \emph{arXiv:2303.09535}, 2023.

\bibitem[Radford et~al.(2021)Radford, Kim, Hallacy, Ramesh, Goh, Agarwal, Sastry, Askell, Mishkin, Clark, Krueger, and Sutskever]{Radford2021LearningTV}
Alec Radford, Jong~Wook Kim, Chris Hallacy, A. Ramesh, Gabriel Goh, Sandhini Agarwal, Girish Sastry, Amanda Askell, Pamela Mishkin, Jack Clark, Gretchen Krueger, and Ilya Sutskever.
\newblock Learning transferable visual models from natural language supervision.
\newblock In \emph{ICML}, 2021.

\bibitem[Ramesh et~al.(2021)Ramesh, Pavlov, Goh, Gray, Voss, Radford, Chen, and Sutskever]{pmlr-v139-ramesh21a}
Aditya Ramesh, Mikhail Pavlov, Gabriel Goh, Scott Gray, Chelsea Voss, Alec Radford, Mark Chen, and Ilya Sutskever.
\newblock Zero-shot text-to-image generation.
\newblock In \emph{Proceedings of the 38th International Conference on Machine Learning}, pages 8821--8831. PMLR, 2021.

\bibitem[Reed et~al.(2016)Reed, Akata, Yan, Logeswaran, Schiele, and Lee]{pmlr-v48-reed16}
Scott Reed, Zeynep Akata, Xinchen Yan, Lajanugen Logeswaran, Bernt Schiele, and Honglak Lee.
\newblock Generative adversarial text to image synthesis.
\newblock In \emph{Proceedings of The 33rd International Conference on Machine Learning}, pages 1060--1069, New York, New York, USA, 2016. PMLR.

\bibitem[Rombach et~al.(2022)Rombach, Blattmann, Lorenz, Esser, and Ommer]{Rombach_2022_CVPR}
Robin Rombach, Andreas Blattmann, Dominik Lorenz, Patrick Esser, and Bj\"orn Ommer.
\newblock High-resolution image synthesis with latent diffusion models.
\newblock In \emph{Proceedings of the IEEE/CVF Conference on Computer Vision and Pattern Recognition (CVPR)}, pages 10684--10695, 2022.

\bibitem[Ruiz et~al.(2023)Ruiz, Li, Jampani, Pritch, Rubinstein, and Aberman]{ruiz2023dreambooth}
Nataniel Ruiz, Yuanzhen Li, Varun Jampani, Yael Pritch, Michael Rubinstein, and Kfir Aberman.
\newblock Dreambooth: Fine tuning text-to-image diffusion models for subject-driven generation.
\newblock In \emph{CVPR}, pages 22500--22510. {IEEE}, 2023.

\bibitem[Salimans and Ho(2022)]{salimans2022progressive}
Tim Salimans and Jonathan Ho.
\newblock Progressive distillation for fast sampling of diffusion models.
\newblock In \emph{International Conference on Learning Representations}, 2022.

\bibitem[Singer et~al.(2023)Singer, Polyak, Hayes, Yin, An, Zhang, Hu, Yang, Ashual, Gafni, Parikh, Gupta, and Taigman]{singer2023makeavideo}
Uriel Singer, Adam Polyak, Thomas Hayes, Xi Yin, Jie An, Songyang Zhang, Qiyuan Hu, Harry Yang, Oron Ashual, Oran Gafni, Devi Parikh, Sonal Gupta, and Yaniv Taigman.
\newblock Make-a-video: Text-to-video generation without text-video data.
\newblock In \emph{The Eleventh International Conference on Learning Representations}, 2023.

\bibitem[Song et~al.(2020)Song, Meng, and Ermon]{song2020denoising}
Jiaming Song, Chenlin Meng, and Stefano Ermon.
\newblock Denoising diffusion implicit models.
\newblock \emph{arXiv:2010.02502}, 2020.

\bibitem[Tang et~al.(2023)Tang, Jia, Wang, Phoo, and Hariharan]{tang2023dift}
Luming Tang, Menglin Jia, Qianqian Wang, Cheng~Perng Phoo, and Bharath Hariharan.
\newblock Emergent correspondence from image diffusion.
\newblock \emph{arXiv preprint arXiv:2306.03881}, 2023.

\bibitem[Truong et~al.(2022)Truong, Danelljan, Yu, and Gool]{denseMatching}
Prune Truong, Martin Danelljan, Fisher Yu, and Luc~Van Gool.
\newblock Probabilistic warp consistency for weakly-supervised semantic correspondences.
\newblock In \emph{{IEEE/CVF} Conference on Computer Vision and Pattern Recognition, {CVPR} 2022, New Orleans, LA, USA, June 18-24, 2022}, pages 8698--8708. {IEEE}, 2022.

\bibitem[Tumanyan et~al.(2023)Tumanyan, Geyer, Bagon, and Dekel]{Tumanyan_2023_CVPR}
Narek Tumanyan, Michal Geyer, Shai Bagon, and Tali Dekel.
\newblock Plug-and-play diffusion features for text-driven image-to-image translation.
\newblock In \emph{Proceedings of the IEEE/CVF Conference on Computer Vision and Pattern Recognition (CVPR)}, pages 1921--1930, 2023.

\bibitem[Wang et~al.(2023)Wang, Yuan, Zhang, Chen, Wang, Zhang, Shen, Zhao, and Zhou]{2023videocomposer}
Xiang Wang, Hangjie Yuan, Shiwei Zhang, Dayou Chen, Jiuniu Wang, Yingya Zhang, Yujun Shen, Deli Zhao, and Jingren Zhou.
\newblock Videocomposer: Compositional video synthesis with motion controllability.
\newblock \emph{CoRR}, abs/2306.02018, 2023.

\bibitem[Wu et~al.(2022)Wu, Ge, Wang, Lei, Gu, Hsu, Shan, Qie, and Shou]{wu2022tuneavideo}
Jay~Zhangjie Wu, Yixiao Ge, Xintao Wang, Stan~Weixian Lei, Yuchao Gu, Wynne Hsu, Ying Shan, Xiaohu Qie, and Mike~Zheng Shou.
\newblock Tune-a-video: One-shot tuning of image diffusion models for text-to-video generation.
\newblock \emph{arXiv preprint arXiv:2212.11565}, 2022.

\bibitem[Xing et~al.(2023)Xing, Xia, Liu, Zhang, Zhang, He, Liu, Chen, Cun, Wang, Shan, and Wong]{xing2023make}
Jinbo Xing, Menghan Xia, Yuxin Liu, Yuechen Zhang, Yong Zhang, Yingqing He, Hanyuan Liu, Haoxin Chen, Xiaodong Cun, Xintao Wang, Ying Shan, and Tien-Tsin Wong.
\newblock Make-your-video: Customized video generation using textual and structural guidance.
\newblock \emph{arXiv preprint arXiv:2306.00943}, 2023.

\bibitem[Xu et~al.(2018)Xu, Zhang, Huang, Zhang, Gan, Huang, and He]{Tao18attngan}
Tao Xu, Pengchuan Zhang, Qiuyuan Huang, Han Zhang, Zhe Gan, Xiaolei Huang, and Xiaodong He.
\newblock Attngan: Fine-grained text to image generation with attentional generative adversarial networks.
\newblock In \emph{CVPR}, pages 1316--1324. Computer Vision Foundation / {IEEE} Computer Society, 2018.

\bibitem[Yu et~al.(2018)Yu, Lin, Yang, Shen, Lu, and Huang]{yu2018free}
Jiahui Yu, Zhe Lin, Jimei Yang, Xiaohui Shen, Xin Lu, and Thomas~S Huang.
\newblock Free-form image inpainting with gated convolution.
\newblock \emph{arXiv preprint arXiv:1806.03589}, 2018.

\bibitem[Zhang et~al.(2023)Zhang, Rao, and Agrawala]{zhang2023adding}
Lvmin Zhang, Anyi Rao, and Maneesh Agrawala.
\newblock Adding conditional control to text-to-image diffusion models, 2023.

\bibitem[Zhao et~al.(2023{\natexlab{a}})Zhao, Wang, Bao, Li, and Zhu]{zhao2023controlvideo}
Min Zhao, Rongzhen Wang, Fan Bao, Chongxuan Li, and Jun Zhu.
\newblock Controlvideo: Adding conditional control for one shot text-to-video editing.
\newblock \emph{arXiv preprint arXiv:2305.17098}, 2023{\natexlab{a}}.

\bibitem[Zhao et~al.(2023{\natexlab{b}})Zhao, Xie, Hong, Li, and Lee]{zhao2023makeaprotagonist}
Yuyang Zhao, Enze Xie, Lanqing Hong, Zhenguo Li, and Gim~Hee Lee.
\newblock Make-a-protagonist: Generic video editing with an ensemble of experts.
\newblock \emph{arXiv preprint arXiv:2305.08850}, 2023{\natexlab{b}}.

\bibitem[{Zhou} et~al.(2022){Zhou}, {Wang}, {Yan}, {Lv}, {Zhu}, and {Feng}]{2022arXiv221111018Z}
Daquan {Zhou}, Weimin {Wang}, Hanshu {Yan}, Weiwei {Lv}, Yizhe {Zhu}, and Jiashi {Feng}.
\newblock {MagicVideo: Efficient Video Generation With Latent Diffusion Models}.
\newblock \emph{arXiv e-prints}, art. arXiv:2211.11018, 2022.

\end{thebibliography}
